\PassOptionsToPackage{table}{xcolor}
\documentclass[10pt,twocolumn,letterpaper]{article}
\usepackage[final]{cvpr} 
\usepackage{graphicx}
\usepackage{tabularx}
\usepackage{mathtools}
\usepackage{graphicx}
\usepackage{amsmath}
\usepackage{amssymb}
\usepackage{booktabs}
\usepackage{dsfont}
\usepackage{multirow}
\usepackage{multicol}
\usepackage{makecell}
\usepackage[usenames,dvipsnames]{xcolor}
\usepackage{tikz}
\usepackage{pgfplots}
\usepackage{pgfkeys}
\usepackage{xcolor}
\usepackage[font=small,labelsep=period]{caption}
\usepackage{enumitem}

\definecolor{bblue}{rgb}{0.0,0.2,0.8}
\usepackage[pagebackref=true,breaklinks=true,colorlinks=true,urlcolor=magenta,bookmarks=false,citecolor=bblue]{hyperref}
\newcolumntype{Y}{>{\centering\arraybackslash}X}
\newcolumntype{R}{>{\raggedleft\arraybackslash}X}
\newcolumntype{L}{>{\raggedright\arraybackslash}X}

\definecolor{lightgray}{rgb}{0.835, 0.835, 0.835}
\definecolor{lightergray}{rgb}{0.935, 0.935, 0.935}
\definecolor{lightgreen}{rgb}{0.85, 1.0, 0.85}

\makeatletter
\def\adl@drawiv#1#2#3{%
        \hskip.5\tabcolsep
        \xleaders#3{#2.5\@tempdimb #1{1}#2.5\@tempdimb}%
                #2\z@ plus1fil minus1fil\relax
        \hskip.5\tabcolsep}
\newcommand{\cdashlinelr}[1]{%
  \noalign{\vskip\aboverulesep
           \global\let\@dashdrawstore\adl@draw
           \global\let\adl@draw\adl@drawiv}
  \cdashline{#1}
  \noalign{\global\let\adl@draw\@dashdrawstore
           \vskip\belowrulesep}}
\makeatother

\makeatletter
\newlength{\qrr@dimen@}
\expandafter\pretocmd\csname tabular*\endcsname{\setlength{\qrr@dimen@}{#1}}{}{}
\newcommand*{\Rowcolor}[2][\tabcolsep]{%
    \ifx\relax#1\relax\else
        \kern-\the\dimexpr#1\relax
    \fi
    \makebox[0pt][l]{%
        \fboxsep=0pt
        \colorbox{#2}{%
            \strut\kern\qrr@dimen@
        }%
    }%
    \ifx\relax#1\relax\else
        \kern\the\dimexpr#1\relax
    \fi
    \ignorespaces
}
\makeatother

\def\addlegendimage{\csname pgfplots@addlegendimage\endcsname}

\usepackage[capitalize]{cleveref}
\crefname{section}{Sec.}{Secs.}
\Crefname{section}{Section}{Sections}
\Crefname{table}{Table}{Tables}
\crefname{table}{Tab.}{Tabs.}

\usepackage{amsmath,amsfonts,bm}
\usepackage{pifont}

\def\eqref#1{equation~\ref{#1}}

\def\1{\bm{1}}

\DeclareMathAlphabet{\mathsfit}{\encodingdefault}{\sfdefault}{m}{sl}
\SetMathAlphabet{\mathsfit}{bold}{\encodingdefault}{\sfdefault}{bx}{n}

\newcommand{\PF}[1]{}

\newcommand{\KY}[1]{}

\newcommand{\MS}[1]{}

\newcommand{\ZD}[1]{}

\newcommand{\YH}[1]{}

\newcommand{\SPE}[1]{}

\newcommand{\WJ}[1]{}

\newcommand{\comment}[1]{}

\newcommand{\bI}{\mathbf{I}}

\newcommand{\bM}{\mathbf{M}}

\newcommand{\bu}{\mathbf{u}}

\newcommand{\bx}{\mathbf{x}}

\def\cvprPaperID{10} %
\def\confName{CVPR}
\def\confYear{2025}
\title{GoTrack: Generic 6DoF Object Pose Refinement and Tracking}

\newcommand{\namesep}{\hspace{0.8em}}
\author{
Van Nguyen Nguyen$^{1*}$\namesep
Christian Forster$^{2}$ \namesep
Sindi Shkodrani$^{2}$ \namesep
Vincent Lepetit$^{1}$ \namesep\\
Bugra Tekin$^{2}$ \namesep
Cem Keskin$^{2}$ \namesep
Tomas Hodan$^{2}$ \vspace{0.5em} \\
{\normalsize $^{1}$LIGM, École des Ponts \namesep $^{2}$Meta Reality Labs} \vspace{0.5em}\\
}

\begin{document}

\definecolor{darkgreen}{RGB}{0,110,0}
\definecolor{darkred}{RGB}{170,0,0}
\def\greencheckmark{\textcolor{darkgreen}{\checkmark}}
\def\redxmark{\textcolor{darkred}{\text{\ding{55}}}}  %

\newcommand{\todo}[1]{\textcolor{red}{{#1}}}

\definecolor{DarkMagenta}{rgb}{0.7, 0.0, 0.7}
\newcommand{\nguyen}[1]{{\color{DarkMagenta}#1}}
\newcommand{\nguyenrmk}[1]{{\color{DarkMagenta} {\bf [VN: #1]}}}
\newcommand{\bugra}[1]{{\color{teal}#1}}
\newcommand{\bugrarmk}[1]{{\color{teal} {\bf [BT: #1]}}}

\definecolor{DarkOrange}{rgb}{1.0, 0.55, 0.0}
\newcommand{\tom}[1]{{\color{DarkOrange}#1}}
\newcommand{\tomrmk}[1]{{\color{DarkOrange} {\bf [TH: #1]}}}

\newcommand\customparagraph[1]{\vspace{0.7em}\noindent\textbf{#1}}
\newcommand{\mytilde}{\raise.17ex\hbox{$\scriptstyle\sim$}}

\maketitle

{\let\thefootnote\relax\footnotetext{*Work done during Nguyen's internship at Meta.}}

\begin{abstract}
We introduce GoTrack, an efficient and accurate CAD-based method for 6DoF object pose refinement and tracking, which can handle diverse objects without any object-specific training. Unlike existing tracking methods that rely solely on an analysis-by-synthesis approach for model-to-frame registration, GoTrack additionally integrates frame-to-frame registration, which saves compute and stabilizes tracking. Both types of registration are realized by optical flow estimation. The model-to-frame registration is noticeably simpler than in existing methods, relying only on standard neural network blocks (a transformer is trained on top of DINOv2) and producing reliable pose confidence scores without a scoring network. For the frame-to-frame registration, which is an easier problem as consecutive video frames are typically nearly identical, we employ a light off-the-shelf optical flow model. We demonstrate that GoTrack can be seamlessly combined with existing coarse pose estimation methods to create a minimal pipeline that reaches state-of-the-art RGB-only results on standard benchmarks for 6DoF object pose estimation and tracking. Our source code and trained models are publicly available at \href{https://github.com/facebookresearch/gotrack}{https://github.com/facebookresearch/gotrack}.

\end{abstract}

\thispagestyle{plain}
\pagestyle{plain}
\section{Introduction}

\begin{figure}[t]
\newlength{\teaserheight}
\setlength\teaserheight{2cm}
\renewcommand{\arraystretch}{0.5} %
\setlength{\tabcolsep}{1pt} %
\begin{center}
\begin{tabular}{cccc}

 \includegraphics[height=\teaserheight]{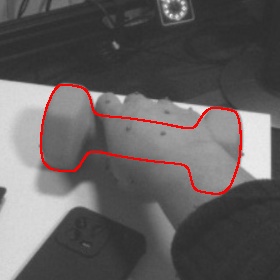} &
 \includegraphics[height=\teaserheight]{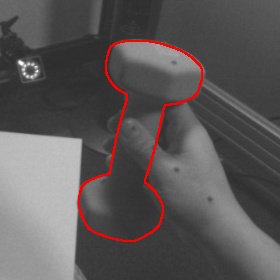} &
 \includegraphics[height=\teaserheight]{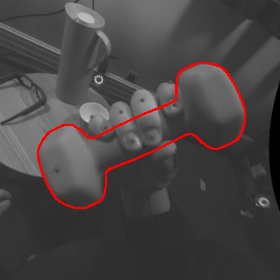} &
 \includegraphics[height=\teaserheight]{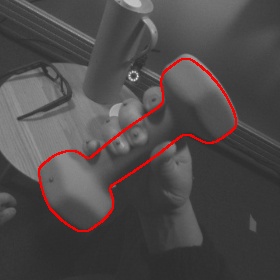}\\

 \includegraphics[height=\teaserheight]{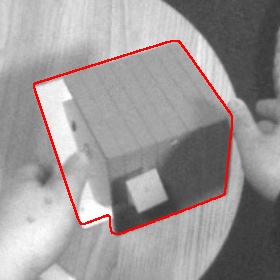} &
 \includegraphics[height=\teaserheight]{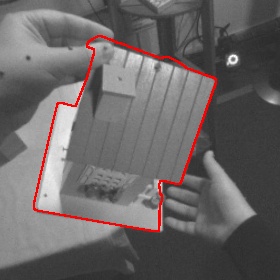} &
 \includegraphics[height=\teaserheight]{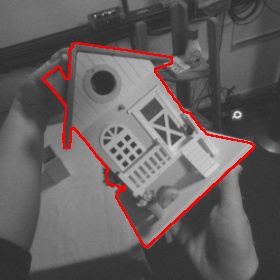} &
 \includegraphics[height=\teaserheight]{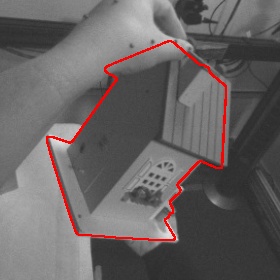}\\

  \includegraphics[height=\teaserheight]{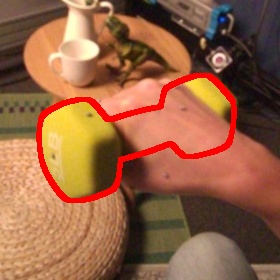} &
 \includegraphics[ height=\teaserheight]{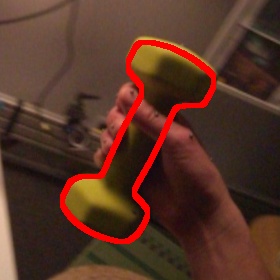} &
 \includegraphics[height=\teaserheight]{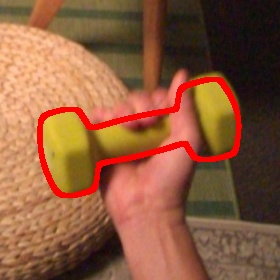} &
 \includegraphics[ height=\teaserheight]{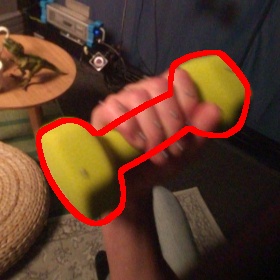}\\

\includegraphics[height=\teaserheight]{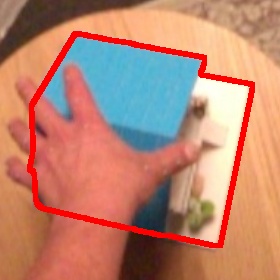} &
 \includegraphics[ height=\teaserheight]{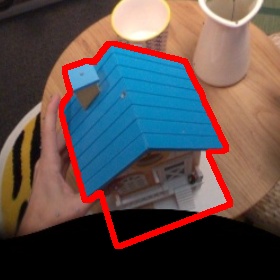} &
 \includegraphics[height=\teaserheight]{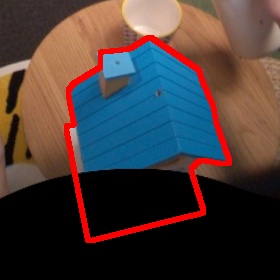} &
 \includegraphics[height=\teaserheight]{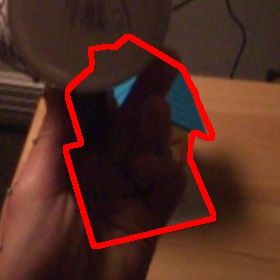}\\

\end{tabular}\\
\begin{tabular}{c}
\includegraphics[ height=0.17\teaserheight]{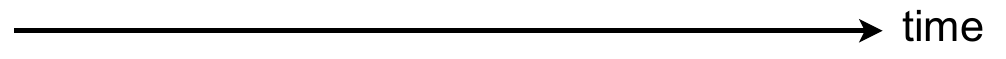}
\end{tabular}
\end{center}
\vspace{-5mm}
\caption{Given a CAD model of the target object, our method can reliably track the 6DoF object pose in challenging conditions such as dynamic scenes with hand-object interactions. The method is applicable to diverse objects and image types, without requiring any domain-specific training.
Each row of this figure shows contours of the tracked model in videos from the HOT3D dataset~\cite{hot3d} captured by Quest 3~\cite{Quest3} and Project Aria~\cite{engel2023project} headsets.
}
\label{fig:teaser}
\end{figure}

Vision-based 6DoF object pose estimation, which is an important task for robotics and AR/VR, has been significantly improved over the past decade~\cite{kehl-iccv17-ssd6d,rad-iccv17-bb8,tekin-cvpr18-realtimeseamlesssingleshot,zakharov-iccv19-dpod,li2018deepim,labbe2020cosypose,Wang_2021_GDRN, liu2022gdrnpp_bop,hodan2020epos,megapose,genflow,nguyen2022template,foundPose,gigaPose}. Yet most methods are designed for single-image input, whereas real-world applications often demand 6DoF object pose tracking, which involves estimating the object pose across consecutive frames.

Classical 6DoF object tracking methods rely on keypoints \cite{Rublee2011, Ozuysal2006, Rosten2005, Skrypnyk2004, Vacchetti2004}, edges \cite{Seo2014, Comport2006, Drummond2002b, Harris1990, stoiber2022srt3d}, or direct optimization \cite{Seo2016, Crivellaro2014, Benhimane2004, Lucas1981, tian2022large}. More recent learning-based tracking methods~\cite{majcher20203d,deng2021poserbpf,wang2023deep,garon2017deep,wen2020se} demonstrate remarkable performance but depend on RGB-D input images~\cite{garon2017deep,wen2020se} or require extensive training datasets specific to each target object~\cite{wang2023deep,deng2021poserbpf,majcher20203d}, which limits their applicability.
A problem that is closely related to pose tracking is pose refinement. In both problems, the input is an image, camera intrinsics, and an initial object pose, and the output is a refined object pose.
Recent methods~\cite{wen2023foundationpose,megapose,genflow} for 6DoF object pose refinement
rely on an analysis-by-synthesis approach, typically rendering a 3D mesh model in the initial pose and predicting a relative pose \wrt the input frame. The pose is predicted either by direct regression~\cite{megapose,wen2023foundationpose} or via 2D-3D correspondences established between the input image and an RGB-D rendering of the object model~\cite{genflow}. In both cases, the methods need to solve the model-to-frame registration, which is not a trivial problem due to the synthetic-to-real domain gap and requires a relatively heavy network.

In simultaneous localization and mapping (SLAM)~\cite{durrant2006simultaneous}, the pose between a 3D scene model and a camera is continuously estimated from video frames via two types of registration: model-to-frame and frame-to-frame. The primary role of model-to-frame registration is to correct drift accumulated over time by re-aligning the current frame with the model, ensuring global consistency. On the other hand, frame-to-frame registration focuses on providing smooth, incremental motion estimation as the camera moves through the environment, allowing SLAM methods to efficiently compute the relative transformation between frames. While the interplay of the two registration types have proven effective for SLAM, the 6DoF object tracking field has been exclusively focused on the model-to-frame registration.

In this paper, we introduce GoTrack, a
method for \underline{g}eneric 6DoF \underline{o}bject \underline{track}ing and pose refinement.
The method adapts an analysis-by-synthesis approach for model-to-frame registration, similarly to recent model-based pose refinement methods, while drawing inspiration from SLAM
and introducing frame-to-frame registration to reduce computational cost when tracking. Both registration types are realized in GoTrack by optical flow estimation.

For the model-to-frame registration, we train a transformer decoder on top of
DINOv2
to predict (1) an optical flow field between a synthetic object template and the input image, and (2) a mask of template pixels that are visible in the input image. From these predictions we construct 2D-3D correspondences by linking visible pixels with 3D points in the model coordinate frame, which are calculated from the depth channel of the template. The 6DoF pose is then calculated by the P\emph{n}P-RANSAC algorithm~\cite{lepetit2009epnp,fischler1981random}.

For the frame-to-frame registration, we build on the observation that this is an easier problem as the differences between consecutive video frames are typically minimal, and solve it with a small off-the-shelf flow estimation network~\cite{teed2020raft}. The frame-to-frame flow is used to propagate 2D-3D correspondences to the next frame, and the more expensive model-to-frame registration is triggered only if the frame-to-frame tracking is lost. Besides saving compute, the use of the frame-to-frame registration is shown to yield more stable tracking results, especially under occlusion.

Additionally, we show how the proposed method for model-to-frame registration can be seamlessly combined with a BoW-based template retrieval from FoundPose~\cite{foundPose} to create an efficient and accurate end-to-end pose estimation pipeline.
We demonstrate that the proposed method reaches state-of-the-art results on the standard 6DoF object pose refinement~\cite{hodan2023bop} and tracking~\cite{xiang2017posecnn,tjaden2018region} benchmarks.

\vspace{1.5ex}
\noindent
In summary, we make the following contributions:
\begin{enumerate}[nosep]
    \item \emph{A flow-based 6DoF pose refiner} that is simpler
    than existing methods, can correct large pose deviations, produces reliable pose confidence score without requiring a scoring network, and reaches state-of-the-art RGB-only results on both 6DoF object pose estimation and tracking benchmarks.

    \item \emph{A seamless 6DoF pose estimation pipeline} that swiftly generates coarse poses by template retrieval and optimizes them by the flow-based refiner. The pipeline is easy to implement
    as both stages rely on
    DINOv2.

    \item \emph{A robust 6DoF pose tracking pipeline} that extends the refiner with frame-to-frame optical flow estimation, which increases tracking efficiency, reduces jitter, and improves robustness to occlusion.

\end{enumerate}

\section{Related work}
In this section, we provide an overview of existing methods for the closely related problems of pose estimation, refinement and tracking of unseen objects.

\customparagraph{Object pose estimation and refinement.} Recent works on 6DoF object pose estimation have introduced diverse benchmarks~\cite{brachmann-eccv14-learning6dobjectposeestimation,hodan-wacv17-tless,Hodan_undated-sl,doumanoglou2016recovering,drost2017introducing,kaskman2019homebreweddb,Xiang2018-dv,sundermeyer2023bop} which have powered many deep learning-based methods~\cite{kehl-iccv17-ssd6d,rad-iccv17-bb8,li2018deepim,tekin-cvpr18-realtimeseamlesssingleshot,li-iccv19-cdpn,zakharov-iccv19-dpod,park-iccv19-pix2pose,hu-cvpr20-singlestage6dobjectposeestimation,labbe2020cosypose,liu2022gdrnpp_bop}.
As shown in the report of the BOP Challenge 2023~\cite{hodan2023bop}, the majority of state-of-the-art methods for 6DoF pose estimation rely on a three-stage pipeline: (1) 2D detection/segmentation, (2) coarse pose estimation, and (3) pose refinement. The top 16 methods for the seen object task and top 13 methods for the unseen object task rely on this pipeline. However, increased accuracy by the third refinement stage usually comes at a significant cost, \eg, the coarse version of GenFlow~\cite{genflow}, the winner of BOP 2023, takes 3.8 seconds per image with around 6 objects on average. In contrast, methods based on rendered templates and estimating coarse poses via nearest neighbor search of frozen features of DINOv2~\cite{oquab2023dinov2}, such as ZS6D~\cite{ausserlechner2023zs6d}, and FoundPose~\cite{foundPose}, have shown promising results while being much faster. Motivated by these results, we build our pose refiner on top of frozen DINOv2~\cite{oquab2023dinov2}, which also enables a seamless integration with the recent coarse pose methods.

The refinement stage, which is crucial for achieving state-of-the-art accuracy, is usually realized by an iterative render-and-compare approach with direct regression (CosyPose\cite{labbe2020cosypose} for seen objects, MegaPose\cite{megapose} and FoundationPose~\cite{wen2023foundationpose} for unseen objects), or with 2D-3D correspondences (GenFlow\cite{genflow}). A common property of these methods is that they require a pose ranking/selection step, which requires an extra network that takes as input the test image and renderings of the object in estimated poses, and outputs pose scores.
In contrast, our approach produces a reliable pose score
without requiring an additional scoring network.

\customparagraph{Object pose tracking.} Classical object tracking methods can be categorized into keypoint-based \cite{Rublee2011, Ozuysal2006, Rosten2005, Skrypnyk2004, Vacchetti2004}, edge-based \cite{Seo2014, Comport2006, Drummond2002b, Harris1990, stoiber2022srt3d}, and direct optimization \cite{Seo2016, Crivellaro2014, Benhimane2004, Lucas1981, tian2022large}. While keypoints and direct optimization are not suitable for texture-less objects, edge-based methods typically struggle with background clutter and texture. To overcome these issues, learning-based methods~\cite{majcher20203d,deng2021poserbpf,wang2023deep,garon2017deep,wen2020se} have been proposed. However, they either depend on RGB-D input images~\cite{garon2017deep,wen2020se} or require extensive training data specific to each target object~\cite{wang2023deep,deng2021poserbpf,majcher20203d}. To address these issues, Nguyen et al.~\cite{nguyen_pizza_2022} proposed a method for tracking 6DoF pose of unseen objects. While their approach only demonstrates results in simple scenarios or simple motions, we focus on more challenging conditions where objects can be heavily occluded or manipulated by hands. Another line of work includes category-level methods~\cite{wang20206,lin2022keypoint}, which focus on unseen instances belonging to training categories. Recently, several methods for instance-level tracking~\cite{wen2020se,li2018deepim,deng2021poserbpf,garon2018framework}, model-based~\cite{stoiber2022iterative,wuthrich2013probabilistic,issac2016depth}, and model-free tracking of unseen objects~\cite{wen2021bundletrack,wen2023bundlesdf} have been proposed. However, these methods either focus on seen objects or use depth images, which limits their applicability.

\customparagraph{Object pose refinement for tracking of unseen objects.}
The most similar to our approach are methods introduced in~\cite{li2018deepim,genflow,megapose,wen2023foundationpose}. A prominent example is DeepIM~\cite{li2018deepim}, which introduced a deep-learning approach for iterative render-and-compare pose refinement. While their method shows generalization to unseen objects, the experiments are limited to synthetic images. MegaPose~\cite{megapose} has further extended this approach and shown strong generalization to unseen objects in real images by training on large-scale datasets. In the BOP Challenge 2023~\cite{hodan2023bop}, GenFlow~\cite{genflow} won the award by introducing a RAFT~\cite{teed2020raft}-based architecture to estimate the flow from the template to real input images. In this work, we also introduce a flow-based method for unseen object pose refinement. However, our approach does not require costly shape constraints, differentiable P\emph{n}P, specialized neural architecture, nor a pose scoring stage to reach state-of-the-art results. Moreover, all of the existing methods focus solely on model-to-frame registration. When applied to tracking, they transition to the next frame solely via pose parameters, discarding pixel-level information about the established registration. In this work, we show that propagating 2D-3D correspondences via frame-to-frame flow increases efficiency and reduces jitter.

\section{Method}

In this section we first propose a 6DoF pose refinement method for unseen objects (Sec.~\ref{sec:refiner}), and then show how this method can support efficient pipelines for estimating accurate object pose from a single image (Sec.~\ref{sec:pose_est_pipeline}) and for tracking the object pose in an image sequence (Sec.~\ref{sec:tracking_pipeline}).

\subsection{Flow-based object pose refinement} \label{sec:refiner}

\noindent\textbf{Problem formulation.}
Given a CAD model of an object, an RGB image with known intrinsics that shows the object in an unknown pose, and a coarse estimate of the object pose, our objective is to refine the pose such as the 2D projection of the model aligns closely with the object’s appearance in the image. We assume the object is rigid and represent its pose by a rigid transformation $(\mathbf{R}, \mathbf{t})$, where $\mathbf{R}$ is a 3D rotation and $\mathbf{t}$ is a 3D translation from the model coordinate frame to the camera coordinate frame.

\customparagraph{Method overview.} We adapt an analysis-by-synthesis approach, similarly to several recent methods for model-based pose refinement~\cite{li2018deepim,megapose,wen2023foundationpose,genflow}. In particular, we predict the refined pose based on (i) an RGB-D template showing the CAD model in the given coarse pose, and (ii) a crop of the object region in the input RGB image (the region is determined by the CAD model and the coarse pose).
Instead of directly regressing a relative 6DoF pose between the rendering and the crop (as in~\cite{li2018deepim,megapose,wen2023foundationpose}), we train a network to predict dense 2D-2D correspondences, represented as an optical flow field,
together with a mask of pixels that are visible in both images. 
The pose is then estimated by lifting the correspondences to 2D-3D using the template depth, and by applying P\emph{n}P-RANSAC~\cite{lepetit2009epnp,fischler1981random}.
The refinement method can be applied in several iterations for increased accuracy.

Predicting 2D outputs (flow and visibility mask) instead of the relative 6DoF pose makes the network easier to train, interpretable and versatile -- we show how to use the predicted correspondences to calculate a reliable and controllable pose score and how to naturally propagate them via frame-to-frame flow in case of tracking. The correspondence-based approach could also support 2D tracking (just by turning off the 3D reasoning part). 

\begin{figure*}[t]
  \centering
   \includegraphics[width=\linewidth]{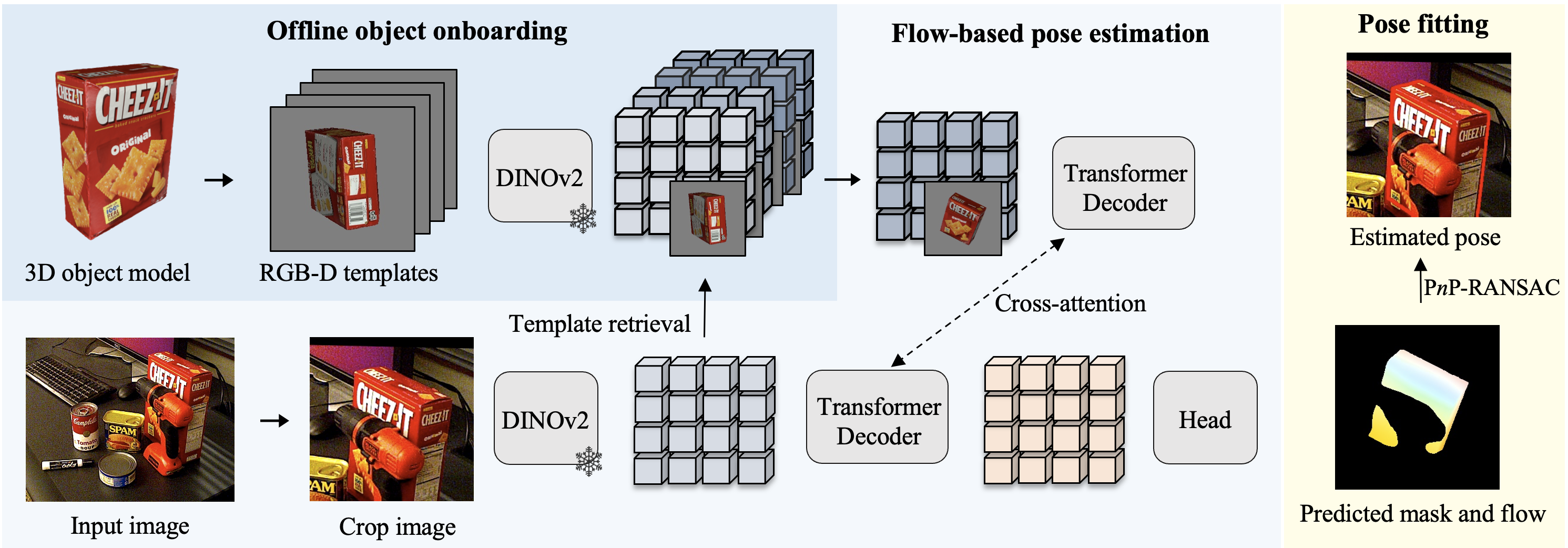}
   \caption{{\bf Overview of the proposed efficient pipeline for 6DoF pose estimation.} Each object is first onboarded by rendering RGB-D templates of its 3D model, and by extracting DINOv2~\cite{oquab2023dinov2} feature maps from RGB channels of the templates.
   At inference, we first crop the input image around a 2D object detection (from CNOS~\cite{nguyen2023cnos}) and retrieve a template that shows the object in a similar pose as in the crop by the BoW-based approach from~\cite{foundPose}. Then we use a transformer decoder~\cite{weinzaepfel2023croco} with a DPT head~\cite{ranftl2021vision} to predict (1) 2D-2D correspondences between the template and the input, and (2) a mask of template pixels that are visible in the input. Next we lift the 2D-2D correspondences established at visible template pixels to 2D-3D using the depth channel of the template, and estimate the pose by P\textit{n}P-RANSAC.}
   \label{fig:refiner}
\end{figure*}

\customparagraph{Predicting 2D-3D correspondences.} Let $\bI_c$ be a perspective crop of the input RGB image around a provided 2D bounding box of the object. The crop is obtained as in~\cite{foundPose} by warping the image to a virtual pinhole camera whose optical axis passes through the center of the 2D bounding box.
Then let $\bI_t$ be an RGB-D template that is rendered using a pinhole camera and shows the object in an initial pose $(\mathbf{R}_0, \mathbf{t}_0)$, let $\bM_t$ be a binary mask of the object silhouette in $\bI_t$, and let $O$ be a set of 3D points representing the surface of the object model. $\bI_t$ and $\bI_c$ have the same resolution.

Given the crop $\bI_c$ and the template $\bI_t$, we aim to establish correspondences between their pixels. We deem pixels $\bu_t \in \bI_t$ and $\bu_c \in \bI_c$ as corresponding if they are aligned with 2D projections of the same 3D point $\mathbf{x} \in O$: $\bu_t = \pi_t(\mathbf{R}_0 \mathbf{x} + \mathbf{t}_0)$ and $\bu_c = \pi_c(\bar{\mathbf{R}} \mathbf{x} + \bar{\mathbf{t}})$, where $\pi$$: \mathbb{R}^3 \mapsto \mathbb{R}^2$ is the 2D projection operator, and $(\bar{\mathbf{R}}, \bar{\mathbf{t}})$ is the GT object pose.

To establish the correspondences, we train a neural network to predict two outputs for each pixel $\bu_t \in \bM_t$: a~likelihood $v_{t \rightarrow c}(\bu_t)$ that the corresponding pixel $\bu_c \in \bI_c$ is visible, and a 2D flow vector $\mathbf{f}_{t \rightarrow c}(\bu_t)$ such that $ \bu_t + \mathbf{f}_{t \rightarrow c}(\bu_t) = \bu_c$.
Given the network predictions, we collect a set of weighted 2D-2D correspondences $D =\{(\bu_c, \bu_t, w)\}$ by linking pixels $\bu_u$ and $\bu_t$ if they are related by $\mathbf{f}_{t \rightarrow c}(\bu_t)$ and if $v_{t \rightarrow c}(\bu_t)$ is above a threshold $\tau_v$. The weight $w$ is defined by $v_{t \rightarrow c}(\bu_t)$. Finally, we convert each linked template pixel $\bu_t$ to a 3D point $\bx \in O$ using the template depth and known camera intrinsics, to obtain a set of weighted 2D-3D correspondences $C =\{(\bu_c, \bx, w)\}$.

\customparagraph{Pose fitting.} The refined pose $(\hat{\mathbf{R}}, \hat{\mathbf{t}})$ is estimated from the established correspondences $C$ by solving the Perspective-\emph{n}-Point (P\emph{n}P) problem. As in~\cite{foundPose}, we solve this problem by the EP\emph{n}P algorithm~\cite{lepetit2009epnp} combined with the RANSAC fitting scheme~\cite{fischler1981random} for robustness. In this scheme, P\emph{n}P is solved repeatedly on a randomly sampled minimal set of 4 correspondences, and the output is defined by the pose hypothesis with the highest number of inlier correspondences $C' \subseteq C$, for which the 2D re-projection error~\cite{lepetit2009epnp} is below a threshold $\tau_r$. The final pose is then refined from all inliers by the Levenberg-Marquardt optimization~\cite{more2006levenberg}.

\customparagraph{Quality of pose estimates.} We define the quality of an estimated pose $(\hat{\mathbf{R}}, \hat{\mathbf{t}})$ as: $q = s'/s$, where $s'$ is the sum of weights $w$ of inlier correspondences $C'$, and $s$ is the sum of weights $w$ of all correspondences $C$. As shown in Sec.~\ref{sec:experiments}, this simple quality measure is surprisingly effective for selecting the best pose estimate in a multiple-hypotheses refinement setup. Note that existing methods~\cite{megapose,genflow,wen2023foundationpose} need a special pose scoring network for this purpose.

\customparagraph{Network architecture and training.} We rely on a frozen DINOv2~\cite{oquab2023dinov2} backbone that we use to extract features from the image crop $\bI_c$ and the template image $\bI_t$. We then pass the features to a two-branch transformer-based decoder from CroCov2~\cite{weinzaepfel2023croco} (one branch for the template and one for the crop), and finally obtain the predictions by applying the DPT head~\cite{ranftl2021vision} to the output of the template branch. The decoder and the head are trained jointly by minimizing the following loss averaged over all pixels $\mathbf{u}_t \in \bM_t$:

\vspace{-12pt}
\begin{align*}
\mathcal{L}(\mathbf{u}_t) = & \text{BCE}\big( v_{t \rightarrow c}(\bu_t), \; \bar{v}_{t \rightarrow c}(\bu_t) \big) \\
& + \bar{v}_{t \rightarrow c}(\bu_t) \; \big\Vert \mathbf{f}_{t \rightarrow c}(\bu_t) - \bar{\mathbf{f}}_{t \rightarrow c}(\bu_t)\big\Vert_1,
\end{align*}
\vspace{-11pt}

\noindent where $\text{BCE}$ is the binary cross entropy loss between the predicted visibility $v_{t \rightarrow c}(\bu_t)$ and the binary ground-truth visibility $\bar{v}_{t \rightarrow c}(\bu_t)$. As in RAFT~\cite{teed2020raft}, the optical flow prediction is supervised by minimizing the L1 distance between the predicted flow vector $\mathbf{f}_{t \rightarrow c}(\bu_t)$ and the ground-truth flow vector $\bar{\mathbf{f}}_{t \rightarrow c}(\bu_t)$. Note that the flow prediction is supervised only at template pixels $\mathbf{u}_t \in \bM_t$ for which the corresponding crop pixel $\mathbf{u}_c \in \bM_c$ is visible, \ie, when $\bar{v}_{t \rightarrow c}(\bu_t) = 1$. 
The ground-truth visibility value $\mathbf{u}_c \in \bM_c$ is defined by $ \bM_c (\bu_t + \mathbf{f}_{t \rightarrow c}(\bu_t))$, where $\bM_c$ is a modal mask defined in the crop ($\bM_c$ is required only at training).

We train the network on the MegaPose-GSO dataset~\cite{megapose} that offers 1M PBR images rendered with BlenderProc~\cite{denninger2020blenderproc}, shows 1000 GSO objects~\cite{downs2022google} that are annotated with the GT 6DoF poses and 2D segmentation masks. As in~\cite{megapose}, we apply heavy image augmentation during training, which avoid overfitting to the synthetic image domain and enables applying the network to real images at inference time. We also randomly perturb the GT poses as in~\cite{megapose}.

\subsection{Efficient object pose estimation pipeline} \label{sec:pose_est_pipeline}

Similarly to the state-of-the-art methods for 6DoF pose estimation of unseen objects~\cite{hodan2023bop}, we adopt a three-stage pipeline: 2D object detection, coarse 6DoF pose estimation, 6DoF pose refinement. While the first stage can be realized by any 2D object detector (\eg CNOS~\cite{nguyen2023cnos}), the key difference of our pipeline is in the other two stages, where we apply efficient template retrieval for coarse pose estimation and our flow-based approach for pose refinement, both based on frozen DINOv2~\cite{oquab2023dinov2}. This streamlined pipeline delivers competitive accuracy while being
faster than existing methods (Sec.~\ref{sec:experiments}). 
Fig.~\ref{fig:refiner} illustrates the pipeline.

\customparagraph{Object onboarding.} First, in an offline onboarding stage, we pre-render a set of RGB-D templates (800 in our experiments), similarly to~\cite{foundPose}. Given a texture-mapped 3D object model, we render templates showing the model under orientations sampled to uniformly cover the SO(3) group of 3D rotations~\cite{alexa2022super}, and the model is rendered using a standard rasterization technique~\cite{shreiner2009opengl} with a gray background and fixed lighting. We also pre-extract DINOv2 feature maps from RGB channels of the templates.

\customparagraph{Coarse pose estimation by template retrieval.} 
Given a 2D object detection (\eg, from CNOS~\cite{nguyen2023cnos}), we generate a coarse object pose with the fast template retrieval approach from FoundPose~\cite{foundPose}.
In Sec.~\ref{sec:experiments}, we show that poses associated with the retrieved templates are sufficient for our flow-based refiner to converge to an accurate pose. 

\customparagraph{Seamless integration.} The template retrieval approach from~\cite{foundPose} leverages Bag-of-Words (BoW) descriptors that are computed from patch features from frozen DINOv2. Since our refiner utilizes the same backbone, the feature extraction process needs to be performed only once per image crop. The extracted DINOv2 features can then be used both to retrieve 
templates and to refine the poses associated. Besides sharing the same backbone, the two stages in fact share the exact same template-based object representation.

\begin{figure}[t]
\definecolor{darkgreen}{RGB}{46,134,95}
\definecolor{darkyellow}{RGB}{128,128,0}

  \centering
   \includegraphics[width=0.95\linewidth]{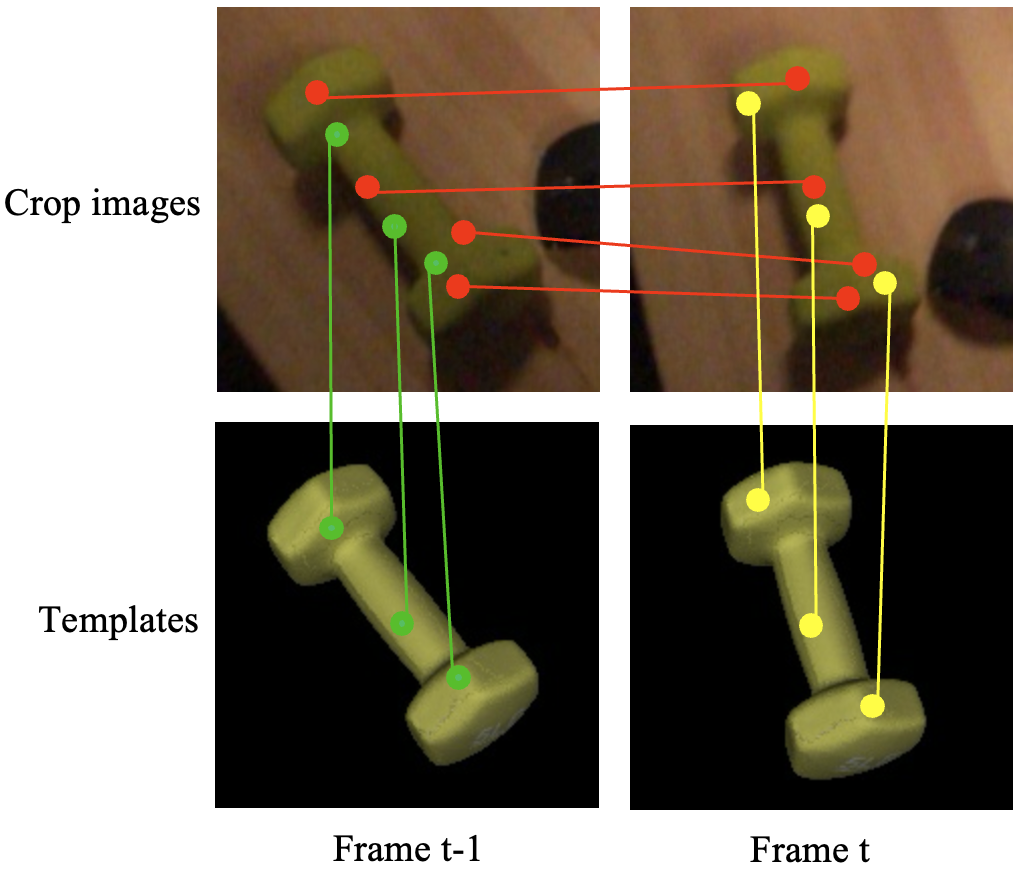}
   \vspace{-5pt}
   \caption{\textbf{Frame-to-frame consistency for tracking.}
   Besides the model-to-frame registration (between a synthetic template and the real input image), which is the main building block of existing 6DoF object tracking methods, we introduce frame-to-frame registration, which reduces jitter and increases efficiency. As the difference between consecutive video frames is typically minimal, estimating frame-to-frame correspondences is a simpler problem and can be solved by a light-weight optical flow network such as RAFT~\cite{teed2020raft}. The model-to-frame correspondences need to be established only in case of a low frame-to-frame tracking confidence.
   }
   \label{fig:teaser}
\end{figure}

\newlength{\plotwidth}
\setlength\plotwidth{2.76cm}
\setlength\lineskip{2.5pt}
\setlength\tabcolsep{2.5pt} 

\begin{figure*}[!t]
\begin{center}
{\small
\begin{tabular}{
>{\centering\arraybackslash}m{\plotwidth}%
>{\centering\arraybackslash}m{\plotwidth}%
>{\centering\arraybackslash}m{\plotwidth}
>{\centering\arraybackslash}m{\plotwidth}%
>{\centering\arraybackslash}m{\plotwidth}%
>{\centering\arraybackslash}m{\plotwidth}
}

Input image & Template  & Predicted mask & Predicted flow & Warped template & Estimated pose \\ 

\frame{\includegraphics[width=\plotwidth,]{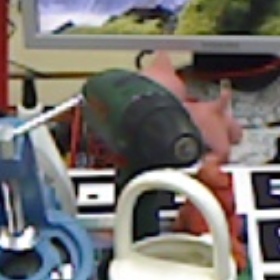}}&
\frame{\includegraphics[width=\plotwidth, ]{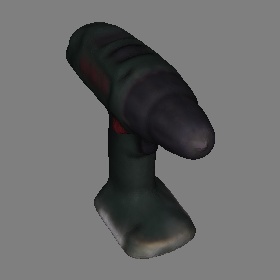}} &
\frame{\includegraphics[width=\plotwidth, ]{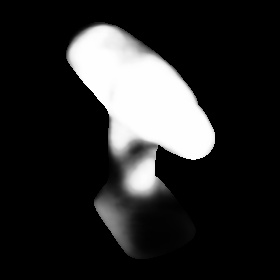}} &
\frame{\includegraphics[width=\plotwidth,]{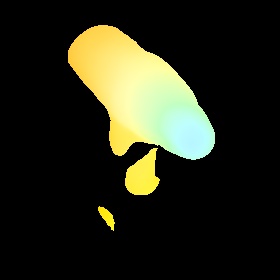}} &
\frame{\includegraphics[width=\plotwidth, ]{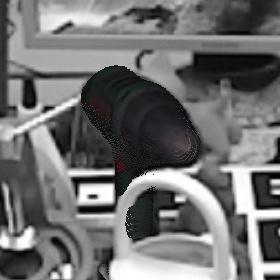}} &
\frame{\includegraphics[width=\plotwidth,]{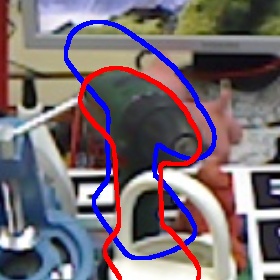}} \\

\frame{\includegraphics[width=\plotwidth,]{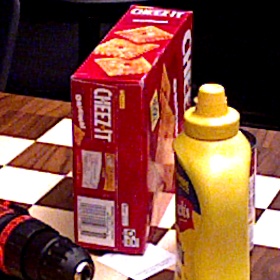}}&
\frame{\includegraphics[width=\plotwidth, ]{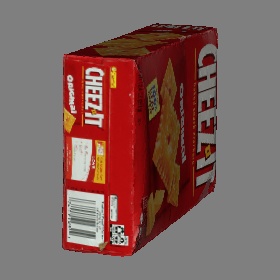}} &
\frame{\includegraphics[width=\plotwidth, ]{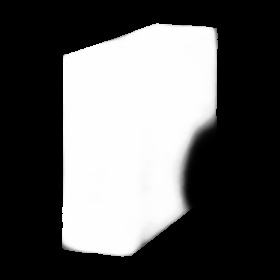}} &
\frame{\includegraphics[width=\plotwidth,]{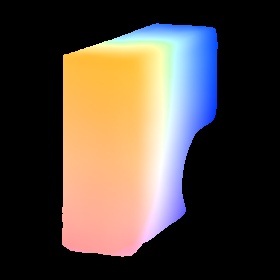}} &
\frame{\includegraphics[width=\plotwidth, ]{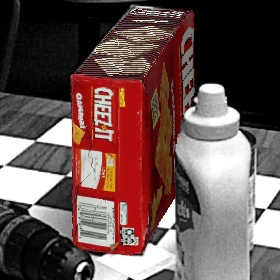}} &
\frame{\includegraphics[width=\plotwidth,]{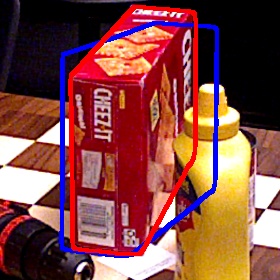}} \\

\frame{\includegraphics[width=\plotwidth,]{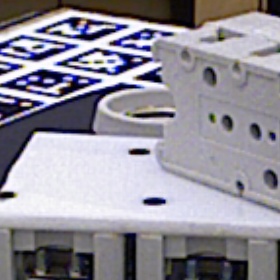}}&
\frame{\includegraphics[width=\plotwidth, ]{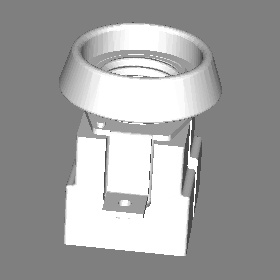}} &
\frame{\includegraphics[width=\plotwidth, ]{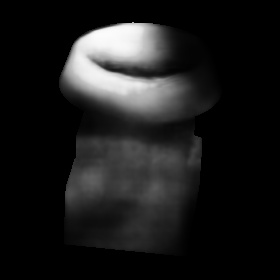}} &
\frame{\includegraphics[width=\plotwidth,]{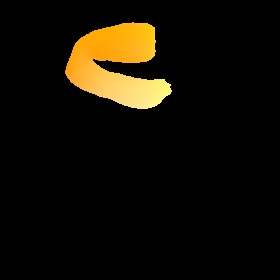}} &
\frame{\includegraphics[width=\plotwidth, ]{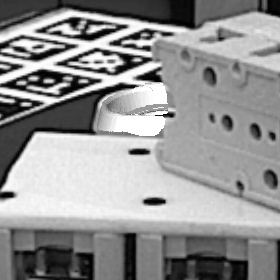}} &
\frame{\includegraphics[width=\plotwidth,]{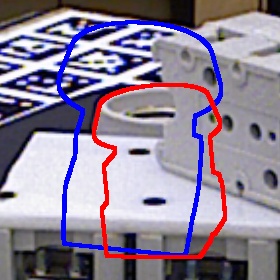}} \\

\end{tabular}
}
    \captionof{figure}{ 
     \textbf{Example results of the GoTrack refiner on LM-O~\cite{brachmann-eccv14-learning6dobjectposeestimation}, YCB-V~\cite{xiang2017posecnn}, and T-LESS~\cite{hodan2017tless}}. The input image is shown in the first column, and the template retrieved using the BoW-based approach~\cite{foundPose} in the second. The third and fourth columns show the predictions of our network, which can then be used to remap pixels from the template to the input image as shown in the fifth column. The last column presents the final pose estimated by P\textit{n}P-RANSAC from 2D-3D correspondences 
     (the contour of the object model in the initial pose is shown in blue, and in the estimated pose in red). As shown in the third column, our method can reliably predict which part of the object is visible, despite never seeing the object during training.
     }
    \label{fig:qualitative}
\end{center}
\end{figure*}

\subsection{Extension to object pose tracking} \label{sec:tracking_pipeline}

\noindent \textbf{Limitation of tracking by refinement.}
The proposed refiner can be directly applied to track the 6DoF object pose in an image sequence. This can be achieved as in \cite{megapose,wen2023foundationpose} by taking the pose from the previous frame as an initial pose for the next frame. Although straightforward, this approach tends to suffers from jitter, \ie, rapid fluctuations or inconsistencies, often resulting in a shaky or unstable appearance in the tracking output. We argue that this limitation arises because predictions between consecutive frames are connected solely through pose parameters. Information about the pixel-level model-to-image registration from the previous frame is discarded. Consequently, the registration process, a core challenge in all analysis-by-synthesis refinement methods, must largely restart in each new frame. We address this limitation by introducing frame-to-frame consistency, which we introduce next.

\customparagraph{Frame-to-frame consistency.} Inspired by  the SLAM literature~\cite{durrant2006simultaneous}, where the pose between a 3D scene model and a camera is continuously estimated from video frames not only by model-to-frame but also by frame-to-frame registration, we introduce the direct frame-to-frame connection to our pipeline by predicting optical flow between the previous and the current video frames.

Specifically, we obtain a perspective crop of the current frame $j$ by constructing a virtual camera looking at the object model in the current pose estimate,
and use RAFT~\cite{teed2020raft} to predict optical flow between the previous and the current crop. We then apply this flow to the 2D coordinates of the inlier 2D-3D correspondences $C'_{i}$ established in the previous frame $i$ to obtain a set of propagated inliers $C'_{i \rightarrow j}$, and re-estimate the pose by the EP\emph{n}P algorithm from $C'_{i \rightarrow j}$. If the inlier ratio \wrt the re-estimated pose is above a threshold $\tau_i$ (typically 0.8), we use the re-estimated pose as the final pose estimate at frame $j$ and proceed to the next frame. Otherwise, if the inlier ratio is below the threshold, we predict the model-to-frame flow (as described earlier) to obtain a new set of 2D-3D correspondences $C_{j}$. In this case, to encourage temporal pose smoothness, we combine sets $C'_{i \rightarrow j}$ and $C_{j}$ by taking all correspondences from the first set and randomly selecting up to $r | C'_{i \rightarrow j} |$ correspondences from the second, \ie, such as the ratio of correspondences selected from the two sets is up to $r$. The pose is then estimated from the combined set by EP\emph{n}P, and inlier correspondences from the combined set are propagated to the next frame.

The parameter $r$, which controls the ratio of correspondences selected from the previous and the current frame, can be
effectively
used to control the trade-off between jitter and the risk of drift (\ie, the risk of deviating from the actual object pose). For lower values of $r$ the tracking tends to be smoother but may suffer from drifting, while for higher values of $r$ the tracking may be jittery.

\section{Experiments} \label{sec:experiments}
In this section, we describe the experimental setup for 6DoF object pose refinement (Sec.~\ref{sec:pose_refinement_exps}) and for 6DoF object pose tracking (Sec.~\ref{sec:pose_tracking_exps}), and compare GoTrack with the state-of-the-art RGB-only methods for these tasks. Specifically, we compare GoTrack with refinement methods MegaPose~\cite{megapose} and GenFlow~\cite{genflow} on BOP datasets~\cite{hodan2023bop}, with tracking methods LDT3D~\cite{tian2022large}, SRT3D~\cite{stoiber2022srt3d} and DeepAC~\cite{wang2023deep} on the RBOT dataset~\cite{tjaden2018region}, and tracking methods PoseCNN~\cite{xiang2017posecnn} and PoseRBPF~\cite{deng2021poserbpf} on the YCB-V dataset~\cite{xiang2017posecnn}. Among the tracking methods, only LDT3D and SRT3D operate on unseen objects while DeepAC, PoseCNN, and PoseRBPF are trained on the test objects.

\newcommand{\ssep}{\hspace{0.7em}}
\begin{table*}[t!]
    \setlength{\tabcolsep}{1.2pt}
    \begin{center}
    \footnotesize
    \begin{tabularx}{1.0\linewidth}{r l YYYYYYY YYYYYYY YY YY }
    \toprule
    
       \multirow{3}{*}{\#\;\;\vspace{0.4ex}}  &
       \multirow{3}{*}{Method} &
       \multicolumn{2}{c}{LM-O}  & 
       \multicolumn{2}{c}{T-LESS}  &
       \multicolumn{2}{c}{TUD-L}  &
       \multicolumn{2}{c}{IC-BIN}  &
       \multicolumn{2}{c}{ITODD}  &
       \multicolumn{2}{c}{HB}  &
       \multicolumn{2}{c}{YCB-V}  &
       \multicolumn{2}{c}{Average$\uparrow$}  &
       \multirow{3}{*}{Time$\downarrow$} & \multirow{3}{*}{Nets$\downarrow$}  \\
      \cmidrule(lr){3-4} \cmidrule(lr){5-6} \cmidrule(lr){7-8} \cmidrule(lr){9-10} \cmidrule(lr){11-12} \cmidrule(lr){13-14} \cmidrule(lr){15-16} \cmidrule(lr){17-18} 
      
        & &\scriptsize{AR$_{\text{MSPD}}$} & \scriptsize{AR} 
        & \scriptsize{AR$_{\text{MSPD}}$} & \scriptsize{AR}
        & \scriptsize{AR$_{\text{MSPD}}$}& \scriptsize{AR} 
        & \scriptsize{AR$_{\text{MSPD}}$}& \scriptsize{AR}
        &\scriptsize{AR$_{\text{MSPD}}$} &\scriptsize{AR}
        &\scriptsize{AR$_{\text{MSPD}}$}& \scriptsize{AR}
        &\scriptsize{AR$_{\text{MSPD}}$} &\scriptsize{AR}
        &\scriptsize{AR$_{\text{MSPD}}$} &\scriptsize{AR} &\\
    \toprule

     \multicolumn{5}{l}{\textit{Coarse pose estimation:}}\\

     \phantom{0}1 &  FoundPose~\cite{foundPose}  & \phantom{0}\textbf{60.7}  &  \textbf{39.6}\phantom{0} &  \phantom{0}\textbf{48.9}  &  \textbf{33.8}\phantom{0} &  \phantom{0}\textbf{67.0}  &  \textbf{46.7}\phantom{0} &  \phantom{0}33.9  &  \textbf{23.9}\phantom{0} &  \phantom{0}\textbf{37.0}  &  \textbf{20.4}\phantom{0}  & \phantom{0}\textbf{65.0}  &  \textbf{50.8}\phantom{0}  &  \phantom{0}\textbf{59.6}  &  \textbf{45.2}\phantom{0} & \phantom{0}\textbf{52.2}  &  \textbf{37.2}\phantom{0}  & $\phantom{0}$1.7 s & 1 \\
     
    \phantom{0}2 &  GigaPose~\cite{gigaPose}  & \phantom{0}51.0  &  29.6\phantom{0} &  \phantom{0}47.9  &  26.4\phantom{0} &  \phantom{0}51.7  &  30.0\phantom{0} &  \phantom{0}\textbf{34.8}  &  22.3\phantom{0} &  \phantom{0}31.4  &  17.5\phantom{0}  & \phantom{0}52.5  &  34.1\phantom{0}  &  \phantom{0}52.5  &  27.8\phantom{0} & \phantom{0}45.4  &  26.8\phantom{0}  & $\phantom{0}$0.4 s & 1 \\

    \phantom{0}3 &  BoW retrieval$^*$  & \phantom{0}25.9 & 11.6\phantom{0}& \phantom{0}28.8 & 15.4\phantom{0}& \phantom{0}25.3 & 13.7\phantom{0}& \phantom{0}17.4 & 10.1\phantom{0}& \phantom{0}31.6 & 17.4\phantom{0}& \phantom{0}55.1 & 42.4\phantom{0}& \phantom{0}15.2 & 11.6\phantom{0} &\phantom{0}28.5 & 17.5\phantom{0} & \phantom{0}\textbf{0.3} s & 1\\

    \midrule
    \multicolumn{8}{l}{\textit{Refinement using a single hypothesis from FoundPose~\cite{foundPose}:}}\\
    
    \Rowcolor{lightergray} \phantom{0}4 & GoTrack  & \phantom{0}\textbf{70.4} &  \textbf{56.5}\phantom{0} & \phantom{0}\textbf{57.3} & 50.4\phantom{0} & \phantom{0}\textbf{81.5} & \textbf{67.2}\phantom{0} & \phantom{0}\textbf{50.7} & \textbf{43.2}\phantom{0} & \phantom{0}\textbf{52.8} & \textbf{39.3}\phantom{0} & \phantom{0}\textbf{76.3} & \textbf{72.2}\phantom{0} & \phantom{0}75.8 & 63.1\phantom{0} & \phantom{0}\textbf{66.4} & \textbf{56.0}\phantom{0} & \phantom{0}\textbf{2.5} s & 1 \\

    \phantom{0}5 & MegaPose~\cite{megapose} & \phantom{0}67.7 & 55.4\phantom{0} & \phantom{0}55.6 & \textbf{51.0}\phantom{0} & \phantom{0}78.4 & 63.3\phantom{0} & \phantom{0}47.4 & 43.0\phantom{0} &  \phantom{0}45.2 & 34.6\phantom{0} & \phantom{0}73.6 & 69.5\phantom{0} & \phantom{0}\textbf{76.0} & \textbf{66.1}\phantom{0} & \phantom{0}63.4 & 54.7\phantom{0} & \phantom{0}4.4 s & 2 \\

    \midrule
    \multicolumn{8}{l}{\textit{Refinement using a single hypothesis from GigaPose~\cite{gigaPose}:}}\\

    \Rowcolor{lightergray} \phantom{0}6  &  GoTrack &  \phantom{0}\textbf{72.6} & 58.4\phantom{0} & \phantom{0}\textbf{61.9} & \textbf{54.9}\phantom{0} & \phantom{0}\textbf{73.9} & \textbf{61.3}\phantom{0} & \phantom{0}\textbf{52.5} & 45.3\phantom{0} & \phantom{0}\textbf{50.0} & 39.5\phantom{0} & \phantom{0}\textbf{77.0} & \textbf{72.9}\phantom{0} & \phantom{0}\textbf{73.8} & 62.0\phantom{0} & \phantom{0}\textbf{66.0} & 56.3\phantom{0} & \phantom{0}\textbf{1.3} s & 2 \\

    \phantom{0}7 &  GenFlow~\cite{genflow}  & \phantom{0}71.1 & \textbf{59.5}\phantom{0} & \phantom{0}60.1 & 55.0\phantom{0} & \phantom{0}72.4 & 60.7\phantom{0} & \phantom{0}52.1 & \textbf{47.8}\phantom{0} & \phantom{0}49.5 & \textbf{41.3}\phantom{0} & \phantom{0}74.7 & 72.2\phantom{0} & \phantom{0}73.2 & 60.8\phantom{0} & \phantom{0}64.7 & \textbf{56.8}\phantom{0} & \phantom{0}2.2 s & 2 \\
    
    \phantom{0}8 & MegaPose~\cite{megapose}  & \phantom{0}68.0 & 55.7\phantom{0} & \phantom{0}59.2 & 54.1\phantom{0} & \phantom{0}71.9 & 58.0\phantom{0} & \phantom{0}49.4 & 45.0\phantom{0} & \phantom{0}45.4 & 37.6\phantom{0} & \phantom{0}73.1 & 69.3\phantom{0} & \phantom{0}73.4 & \textbf{63.2}\phantom{0} & \phantom{0}64.2 & 54.7\phantom{0} & \phantom{0}2.3 s & 2 \\

    \midrule
    \multicolumn{10}{l}{\textit{Refinement using BoW retrieval for coarse pose estimation (from row 3):}} \\

    \Rowcolor{lightergray} \phantom{0}9  &  GoTrack & \phantom{0}70.4 & 56.9\phantom{0} & \phantom{0}56.9 & 50.4\phantom{0} & \phantom{0}79.2 & 65.2\phantom{0} & \phantom{0}49.7 & 42.4 \phantom{0} & \phantom{0}52.6 & 40.2\phantom{0} & \phantom{0}72.5 & 69.5\phantom{0} & \phantom{0}75.2 & 62.9\phantom{0} & \phantom{0}65.2 & 55.4\phantom{0} & 0.8 s & 1\\
    
    \midrule

    \multicolumn{8}{l}{\textit{Pose refinement using 5 hypotheses from GigaPose~\cite{gigaPose}:}}\\
    
    \Rowcolor{lightergray} 12 & GoTrack & \phantom{0}\textbf{78.1} & 62.7\phantom{0} & \phantom{0}\textbf{66.3} & \textbf{59.5}\phantom{0} & \phantom{0}\textbf{80.8} & \textbf{66.8}\phantom{0} & \phantom{0}54.4 & 46.6\phantom{0} & \phantom{0}\textbf{56.3} & 43.0\phantom{0} & \phantom{0}\textbf{81.9} & \textbf{77.1}\phantom{0} & \phantom{0}\textbf{80.6} & \textbf{68.4}\phantom{0} & \phantom{0}\textbf{71.2} & \textbf{60.6}\phantom{0} & \phantom{0}\textbf{3.0} s & 2 \\

    13 &  GenFlow~\cite{genflow}   & \phantom{0}75.3 & \textbf{63.1}\phantom{0} & \phantom{0}63.7 & 58.2\phantom{0} & \phantom{0}79.4 & 66.4\phantom{0} & \phantom{0}\textbf{55.1} & \textbf{49.8}\phantom{0} & \phantom{0}55.1 & \textbf{45.3}\phantom{0} & \phantom{0}78.3 & 75.6\phantom{0} & \phantom{0}78.8 & 65.2\phantom{0} & \phantom{0}69.4 & 60.5\phantom{0} & 10.6 s & 3  \\
    
    14 & MegaPose~\cite{megapose}  & \phantom{0}73.6 & 59.8\phantom{0} & \phantom{0}62.0 & 56.5\phantom{0}  & \phantom{0}77.9 & 63.1\phantom{0} & \phantom{0}52.8 & 47.3\phantom{0} & \phantom{0}48.5 & 39.7\phantom{0} & \phantom{0}76.5 & 72.2\phantom{0} & \phantom{0}77.7 & 66.1\phantom{0} & \phantom{0}65.3 & 57.8\phantom{0}  & \phantom{0}7.7 s & 3 \\

    \bottomrule
    \end{tabularx}
    \end{center}
    \vspace{-4pt}
    \caption{\textbf{Pose refinement performance on the seven core BOP datasets~\cite{hodan2023bop}.} Reported are accuracy scores, the average time required to estimate poses of all objects in an image (in seconds), and the number of different neural networks required by the pipelines.
    }
    \label{tab:main_results}
\end{table*}

\customparagraph{Implementation details.} The evaluated version of the proposed refinement method used 800 templates (with approximately $25^{\circ}$ angle between depicted object orientations), the template and crop resolution of 280$\times$280\,px, DINOv2~\cite{oquab2023dinov2} with the ViT-S architecture, the transformer decoder from CroCov2~\cite{weinzaepfel2023croco} with 12 blocks, the default DPT head~\cite{ranftl2021vision}, the visibility threshold $\tau_v$ set to 0.3, up to 400 iterations of P\emph{n}P-RANSAC, and the re-projection threshold $\tau_r$ set to 4\,px. For the pose refinement experiments, we use 5 refinement iterations as previous works~\cite{megapose,genflow}. Although our flow-based refiner could rely solely on templates pre-rendered in the onboarding phase (at each refinement iteration, we could use the template that best matches the orientation of the current pose estimate), we found that this approach does not perform as well as online rendering, where the 3D model is rendered in the current pose estimate and therefore does not suffer from the pose quantization error.
Online rendering is used also in the prior works~\cite{megapose,genflow}.
For the object tracking experiments,
the parameter $\tau_i$ that triggers the template-to-frame registration was set to 0.8, and the mixing ratio $r$ was set to 2. We trained the flow-based refiner until convergence on the MegaPose-GSO dataset~\cite{megapose} using Adam~\cite{kingma2014adam}, with DINOv2 weights being frozen.

\subsection{6DoF object pose refinement}
\label{sec:pose_refinement_exps}

\customparagraph{Evaluation datasets.}
We evaluate our method on the seven core BOP datasets~\cite{hodan2023bop}: LM-O~\cite{brachmann-eccv14-learning6dobjectposeestimation}, T-LESS~\cite{hodan-wacv17-tless}, TUD-L~\cite{Hodan_undated-sl}, IC-BIN~\cite{doumanoglou2016recovering}, ITODD~\cite{drost2017introducing}, ~\cite{kaskman2019homebreweddb}, and YCB-V~\cite{Xiang2018-dv}. These datasets include the total of 132 different objects and 19048 annotated test object instances.

\customparagraph{Evaluation metrics.} We use the BOP evaluation protocol~\cite{hodan-eccv20-bopchallenge2020on}, which relies on three metrics: Visible Surface Discrepancy~(VSD), Maximum Symmetry-Aware Surface Distance~(MSSD), and Maximum Symmetry-Aware Projection Distance~(MSPD). The final average recall~(AR), is calculated by averaging the individual AR scores of these three metrics across a range of error thresholds. 

\customparagraph{Baselines.} We compare our method with MegaPose~\cite{megapose} and GenFlow~\cite{genflow}, which are currently the state-of-the-art methods on the BOP benchmark. The results of MegaPose and GenFlow are sourced from the BOP leaderboard~\cite{hodan2023bop}.

\customparagraph{Results.} When starting from coarse poses estimated by FoundPose~\cite{foundPose}, our refiner outperforms the MegaPose refiner~\cite{megapose} on the majority of BOP datasets, achieving +3.0 AR$_{\text{MSPD}}$ and +1.3 AR on average (rows 4 and 5 in Tab.~\ref{tab:main_results}). When starting from coarse poses estimated by GigaPose~\cite{gigaPose}, our refiner outperforms MegaPose by similar margins (rows 6 and 8). Compared to GenFlow~\cite{genflow}, our refiner achieves +1.3 AR$_{\text{MSPD}}$ and -0.5 AR (rows 6 and 7). 

In terms of complexity of the model architecture, the simplest from the evaluated methods is the pipeline proposed in Sec.~\ref{sec:pose_est_pipeline}, which relies on the BoW-based template retrieval for coarse pose estimation followed by our refiner. This pipeline relies on only a single neural network (column ``Nets'') that is shared by both stages, which makes it faster and easier to deploy on mobile devices. 
Comparing runtime of other methods is problematic as each was evaluated on a different GPU (we used V100, while MegaPose was evaluated on RTX 2080 and GenFlow on RTX 3090). However, the difference in complexity of the methods is clear. Both GenFlow and MegaPose pipelines rely on different networks for coarse pose estimation and for refinement. In case of multi-hypotheses setup, where multiple coarse poses are considered per object instance and their refined versions need to be at the end ranked to select the final output, the GenFlow and MegaPose pipelines need an additional scoring network, whereas our method uses reliable pose confidence score can be obtained by simply calculating the weighted inlier ratio defined in Sec.~\ref{sec:refiner}.

\begin{table}[t!]
\setlength{\tabcolsep}{0.75pt}
\setlength{\tabcolsep}{1.0pt}
\scriptsize
\begin{center}
\begin{tabularx}{1.0\linewidth}{l YY YY YY YY}

\toprule

 & \multicolumn{2}{c}{PoseCNN~\cite{xiang2017posecnn}} & \multicolumn{2}{c}{PoseRBPF~\cite{deng2021poserbpf}} & \multicolumn{2}{c}{GoTrack}  & \multicolumn{2}{c}{GoTrack+f2f} \\

 \midrule

Initialization  & \multicolumn{2}{c}{-}  & \multicolumn{2}{c}{PoseCNN}   &  \multicolumn{2}{c}{FoundPose}  & \multicolumn{2}{c}{FoundPose} \\

\midrule

Unseen objects  & \multicolumn{2}{c}{\redxmark}  & \multicolumn{2}{c}{\redxmark}     &  \multicolumn{2}{c}{\greencheckmark}  & \multicolumn{2}{c}{\greencheckmark} \\
\toprule

 & {\tiny ADD} & {\tiny ADD-S}  & {\tiny ADD} & {\tiny ADD-S}  & {\tiny ADD}  &  {\tiny ADD-S} & {\tiny ADD} & {\tiny ADD-S} \\

\midrule

\multicolumn{1}{l}{\tiny 002\_master\_chef\_can} & 50.9 & 84.0  & 58.0 & 77.1   & \cellcolor{lightergray}71.0 & \cellcolor{lightergray}\textbf{86.3} &  \cellcolor{lightergray}\textbf{71.3} & \cellcolor{lightergray}\textbf{86.3}\\

\multicolumn{1}{l}{\tiny 003\_cracker\_box} & 51.7 & 76.9  & 76.8 &  87.0   & \cellcolor{lightergray}79.6 & \cellcolor{lightergray}89.2 & \cellcolor{lightergray}\textbf{80.6} & \cellcolor{lightergray}\textbf{89.8}\\

\multicolumn{1}{l}{\tiny 004\_sugar\_box} & 68.6 & 84.3  & 75.9 &  87.6  & \cellcolor{lightergray}80.4 & \cellcolor{lightergray}89.9 & \cellcolor{lightergray}\textbf{80.5} & \cellcolor{lightergray}\textbf{90.0}\\

\multicolumn{1}{l}{\tiny 005\_tomato\_soup\_can}  & 66.0 & 80.9  & \textbf{74.9}  & 84.5  & \cellcolor{lightergray}74.0 & \cellcolor{lightergray}\textbf{87.2} & \cellcolor{lightergray}71.2 & \cellcolor{lightergray}85.8\\

\multicolumn{1}{l}{\tiny 006\_mustard\_bottle}  & 79.9 & 90.2  & 82.5 & 91.0   & \cellcolor{lightergray}83.7 & \cellcolor{lightergray}91.9 & \cellcolor{lightergray}\textbf{84.4} & \cellcolor{lightergray}\textbf{92.2}\\

\multicolumn{1}{l}{\tiny 007\_tuna\_fish\_can} & 70.4 & 87.9 & 59.0 &  79.0   & \cellcolor{lightergray}81.4 & \cellcolor{lightergray}91.7 & \cellcolor{lightergray}\textbf{82.0} & \cellcolor{lightergray}\textbf{91.9}\\

\multicolumn{1}{l}{\tiny 008\_pudding\_box}& 62.9 & 79.0  & 57.2 & 72.1  & \cellcolor{lightergray}82.8 & \cellcolor{lightergray}90.4 &  \cellcolor{lightergray}\textbf{83.5} & \cellcolor{lightergray}\textbf{90.7} \\

\multicolumn{1}{l}{\tiny 009\_gelatin\_box} & 75.2 & 87.1 & \textbf{88.8}  & \textbf{93.1}  & \cellcolor{lightergray}83.4 & \cellcolor{lightergray}90.9 & \cellcolor{lightergray}\textbf{84.4} & \cellcolor{lightergray}\textbf{91.5}\\

\multicolumn{1}{l}{\tiny 010\_potted\_meat\_can}  & 59.6 & 78.5  & 49.3 & 62.0   & \cellcolor{lightergray}80.5 & \cellcolor{lightergray}91.3 & \cellcolor{lightergray}\textbf{83.5} & \cellcolor{lightergray}\textbf{92.8}\\

\multicolumn{1}{l}{\tiny 011\_banana}  & 72.3 & 85.9  & 24.8 & 61.5 & \cellcolor{lightergray}53.8 & \cellcolor{lightergray}\textbf{71.0} & \cellcolor{lightergray}\textbf{56.0} & \cellcolor{lightergray}68.7\\
 
 \multicolumn{1}{l}{\tiny 019\_pitcher\_base}  & 52.5 & 76.8  & 75.3 & 88.4 & \cellcolor{lightergray}79.7 & \cellcolor{lightergray}90.0 &  \cellcolor{lightergray}\textbf{79.9} & \cellcolor{lightergray}\textbf{90.1}\\

\multicolumn{1}{l}{\tiny 021\_bleach\_cleanser}  & 50.5 & 71.9  & 54.5 & 69.3  & \cellcolor{lightergray}68.9 & \cellcolor{lightergray}83.1 & \cellcolor{lightergray}\textbf{69.3} & \cellcolor{lightergray}\textbf{83.3} \\

\multicolumn{1}{l}{\tiny 024\_bowl}  & 6.5 & 69.7 & \textbf{36.1} & \textbf{86.0}  & \cellcolor{lightergray}35.1 & \cellcolor{lightergray}66.2 & \cellcolor{lightergray}34.8 & \cellcolor{lightergray}62.0 \\

\multicolumn{1}{l}{\tiny 025\_mug}  & 57.7 & 78.0 & \textbf{70.9} & 85.4  & \cellcolor{lightergray}70.1 & \cellcolor{lightergray}87.1 & \cellcolor{lightergray}50.8 & \cellcolor{lightergray}\textbf{87.4}\\

\multicolumn{1}{l}{\tiny 035\_power\_drill} &  55.1 & 72.8 & 70.9 & 85.0 & \cellcolor{lightergray}81.9 & \cellcolor{lightergray}91.2 & \cellcolor{lightergray}\textbf{82.1} & \cellcolor{lightergray}\textbf{91.3}\\

\multicolumn{1}{l}{\tiny 036\_wood\_block}  & \textbf{31.8} &  \textbf{65.8}  & 2.8 & 33.3 & \cellcolor{lightergray}0.7 & \cellcolor{lightergray}30.4 & \cellcolor{lightergray}5.3 & \cellcolor{lightergray}40.9\\

\multicolumn{1}{l}{\tiny 037\_scissors} & 35.8 & 56.2  & 21.7 &  33.0 & \cellcolor{lightergray}58.3 & \cellcolor{lightergray}76.5 & \cellcolor{lightergray}\textbf{59.7} & \cellcolor{lightergray}\textbf{77.8}\\

\multicolumn{1}{l}{\tiny 040\_large\_marker} & 58.0 & 71.4  & 48.7 & 59.3  & \cellcolor{lightergray}60.0 & \cellcolor{lightergray}67.3 & \cellcolor{lightergray}\textbf{65.0} & \cellcolor{lightergray}\textbf{73.9}\\

\multicolumn{1}{l}{\tiny 051\_large\_clamp}  & 25.0 & 49.9   & 47.3 &  76.9  & \cellcolor{lightergray}64.1 & \cellcolor{lightergray}83.1 & \cellcolor{lightergray}\textbf{64.4} & \cellcolor{lightergray}\textbf{83.2}\\

\multicolumn{1}{l}{\tiny 052\_extra\_large\_clamp}  & 15.8 & 47.0   & 26.5 & 69.5 & \cellcolor{lightergray}\textbf{83.5} & \cellcolor{lightergray}\textbf{93.5}  & \cellcolor{lightergray}82.8 & \cellcolor{lightergray}93.0 \\

 \multicolumn{1}{l}{\tiny 061\_foam\_brick} & 40.4 &  87.8   & 78.2 & 89.7 & \cellcolor{lightergray}83.0 & \cellcolor{lightergray}92.3 &  \cellcolor{lightergray}\textbf{83.3} & \cellcolor{lightergray}\textbf{92.4} \\

\midrule

 Average & 53.7  & 75.9  & 59.9 & 77.5  & \cellcolor{lightergray}\textbf{69.3} & \cellcolor{lightergray}82.9 & \cellcolor{lightergray}\textbf{69.3} & \cellcolor{lightergray}\textbf{83.6}\\

\bottomrule
 
\end{tabularx}
\end{center}
\caption{\textbf{Tracking performance on the YCB-V dataset~\cite{xiang2017posecnn}.} The accuracy is measured by AUC \wrt the ADD and ADD-S pose error functions. Despite not being trained on the YCB-V objects, our method outperforms PoseCNN~\cite{xiang2017posecnn} and PoseRBPF~\cite{deng2021poserbpf} which were trained on these specific objects. Adding the frame-to-frame consistency (f2f) makes our method 6X more efficient while slightly improving the accuracy (see text for details).} \label{tab:tracking_results}
\end{table}

\begin{table}[t!]
\setlength{\tabcolsep}{1pt} %
\renewcommand{\arraystretch}{1} %
\scriptsize
\begin{center}
\begin{tabularx}{1.0\linewidth}{l c c c c c c c c c c }
\toprule
 \multirow{2}{*}{Method} & \multicolumn{6}{c}{with reset} & \multicolumn{2}{c}{w/o reset} \\
 \cmidrule(lr){2-7} \cmidrule(lr){8-10}
  & \;Regular\; & \;Dynamic\; & \;Noisy\; & \;Occlusion\; & \;\;Mean\;\; & Resets~$\downarrow$\; & \multicolumn{2}{c}{Mean} \\
 
 \midrule

SRT3D~\cite{stoiber2022srt3d} & 94.2 & 94.6 & 81.7 & 93.2 & 90.9 & 6575 & \multicolumn{2}{c}{19.0}\\
LDT3D~\cite{tian2022large} & 95.2 & 95.4 & 83.2 & 94.9 & 92.1 & 6228 & \multicolumn{2}{c}{18.8}\\
DeepAC~\cite{wang2023deep}* & 95.6 & 95.6 & 88.0 &  94.0 &  93.3 & 4826 & \multicolumn{2}{c}{30.3}\\
GoTrack & \bf 97.3 & \bf 96.2 & \bf 94.2 & \bf 95.4 & \bf 95.9 & \bf 3021 & \multicolumn{2}{c}{\bf 86.7} \\
\bottomrule
\end{tabularx}
\end{center}

\caption{\textbf{Tracking on the RBOT dataset~\cite{tjaden2018region} with/without reset} as in~\cite{wang2023deep,tian2022large,stoiber2022srt3d}. The accuracy is measured by 5\textit{cm}-5\textdegree score and the number of reset. Character * indicates that DeepAC is a supervised method trained on the test objects. GoTrack significantly outperforms existing methods across all settings and metrics.} \label{tab:rbot_reset}
\vspace{-1mm}
\end{table}

\subsection{6DoF object pose tracking}
\label{sec:pose_tracking_exps}

\customparagraph{Evaluation datasets.} We evaluate our tracking method on two popular datasets: YCB-V~\cite{xiang2017posecnn} and RBOT~\cite{tjaden2018region}.
Additional tracking results on the recent HOT3D~\cite{hot3d} dataset are presented in the appendix.

\customparagraph{Evaluation metrics.} We use the same standard evaluation metrics as in~\cite{xiang2017posecnn,deng2021poserbpf,wang2023deep,stoiber2022srt3d,tian2022large}: Area Under the Curve (AUC) for the ADD and and ADD(-S) pose error functions on YCB-V dataset, and 5\textit{cm}-5\textdegree and the number of resets on RBOT dataset. A reset is counted when the tracking method fails, \ie, when the translation error exceeds 5 centimeters or the rotation error exceeds 5 degrees, the method is re-initialized with the ground-truth pose.
More details about these metrics are in the appendix.

\customparagraph{Baselines.} Since no RGB-only methods for tracking of unseen objects have been evaluated on the YCB-V dataset, we compare our method with tracking methods for seen objects (\ie, methods trained on the target objects), PoseCNN~\cite{xiang2017posecnn} and PoseRBPF~\cite{deng2021poserbpf}. PoseCNN is a method for 6DoF object pose estimation from a single image and was evaluated frame by frame. PoseRBPF was initialized at the first frame with a pose estimated by PoseCNN. Since the PoseCNN estimates are not publicly available, we initialized our tracking method with a pose estimate from FoundPose. On the RBOT dataset, we compare our method with contour-based methods DeepAC~\cite{wang2023deep} and SRT3D~\cite{stoiber2022srt3d} and the optimization-based method LDT3D~\cite{tian2022large}.

\customparagraph{Results on YCB-V~\cite{xiang2017posecnn}.} As shown in Tab.~\ref{tab:tracking_results}, even though our method was not trained on the YCB-V objects, it significantly outperforms both PoseCNN~\cite{xiang2017posecnn} and PoseRBPF~\cite{deng2021poserbpf}, which were trained specifically for these objects. Adding the frame-to-frame flow makes our method around 6X more efficient while slightly improving the accuracy. The forward pass of the flow-based refiner, which we use for the template-to-frame registration, takes 40\,ms while the forward pass of RAFT-Small, which we use for the frame-to-frame registration takes only 6\,ms. The template-to-frame flow is triggered only every 50th frame on average on sequences from the YCB-V dataset. 
Another effect of adding the frame-to-frame flow is a reduced jitter, which can be seen in the supplementary video (App.~\ref{app:track_qual}).

\customparagraph{Results on RBOT~\cite{tjaden2018region}.} As shown in Tab. \ref{tab:rbot_reset}, GoTrack outperforms LDT3D, SRT3D and DeepAC on all RBOT sequence types while requiring 37--54\% less resets. The most noticeable difference is on the noisy sequences where contour detection is less reliable. Furthermore, these methods work well only on objects with distinct contour -- they struggle if the same contour shape corresponds to multiple different poses and the actual pose can be estimated only based on texture (\eg, DeepAC achieves only 70.1\% 5\textit{cm}-5\textdegree accuracy on the textured and cylindrical soda object from RBOT while GoTrack achieves 97.2\%). Additionally, Tab.~\ref{tab:rbot_reset} shows that our method significantly outperforms the other methods in tracking without reset (86.7 vs 30.3 AUC).

\section{Conclusion}

We introduced a method for 6DoF pose refinement of unseen objects, based on predicting template-to-frame optical flow along with a mask of visible template pixels. We showed how our refinement method can be seamlessly combined with a BoW-based template retrieval to create an efficient and accurate object pose estimation pipeline. Additionally, we proposed an extension for 6DoF pose tracking of unseen objects by propagating 2D-3D correspondences using frame-to-frame optical flow. Our approach achieves state-of-the-art results on the standard 6DoF object pose refinement and tracking benchmarks.
For future work, we plan to draw further inspiration from SLAM literature and explore extending our approach to a model-free setup.

{\small
\bibliographystyle{ieee_fullname}
\bibliography{cleaned_refs}
}

\clearpage
\appendix
\section*{Appendix}
In this appendix, we provide details about the evaluation metrics used in the object tracking experiments (Sec.~\ref{app:appendix_pose_tracking}). We also provide quantitative tracking results on the HOT3D~\cite{hot3d} dataset (Sec.~\ref{app:track_quant}), and additional qualitative results for both refinement and tracking (Sec.~\ref{app:ref_qual} and~\ref{app:track_qual}). 

\section{Evaluation metrics for tracking}\label{app:appendix_pose_tracking}
As discussed in Section 4.2 of the main paper, we use the same standard evaluation metrics as previous works~\cite{xiang2017posecnn,deng2021poserbpf,wang2023deep,stoiber2022srt3d,tian2022large}: Area under the Curve (AUC) for the ADD and and ADD(-S) pose error functions on YCB-V dataset, and 5\textit{cm}-5\textdegree, and number of resets on RBOT dataset. The ADD error~\cite{hinterstoisser-accv12-modelbasedtrainingdetection} measures the average deviation of the model vertices in the estimated pose. The ADD-S error measures the average distance to the closest model point~\cite{hinterstoisser-accv12-modelbasedtrainingdetection,hodan2016evaluation} as is suitable for evaluating pose estimates of symmetric objects. The AUC values are calculated by averaging the recall rates \wrt ADD and ADD-S for multiple thresholds ranging from 0.1\,cm to 10\,cm. The 5\textit{cm}-5\textdegree score refers to the success rate of tracking poses with a translation error of less than 5 centimeters and a rotation error of less than 5 degrees. On RBOT dataset, after the initialization of the tracker with the GT pose, the tracker runs until either the recorded sequence ends or tracking was unsuccessful. In the case of unsuccessful tracking, the algorithm is re-initialized with the GT pose, and we count the total number of re-initialization.

\section{Quantitative tracking results} \label{app:track_quant}

Tab.~\ref{tab:hot3d} shows results on the HOT3D~\cite{hot3d} recordings from Aria and Quest 3.
Since the ground-truth annotations for the test set of the HOT3D dataset are not publicly available, we evaluate GoTrack on the training set of HOT3D. Despite not being trained on grayscale images, our method achieves a similar performance on grayscale images from HOT3D-Quest3 as on RGB images from HOT3D-Aria.

\begin{figure*}[!h]
  \centering
   \includegraphics[width=\linewidth]{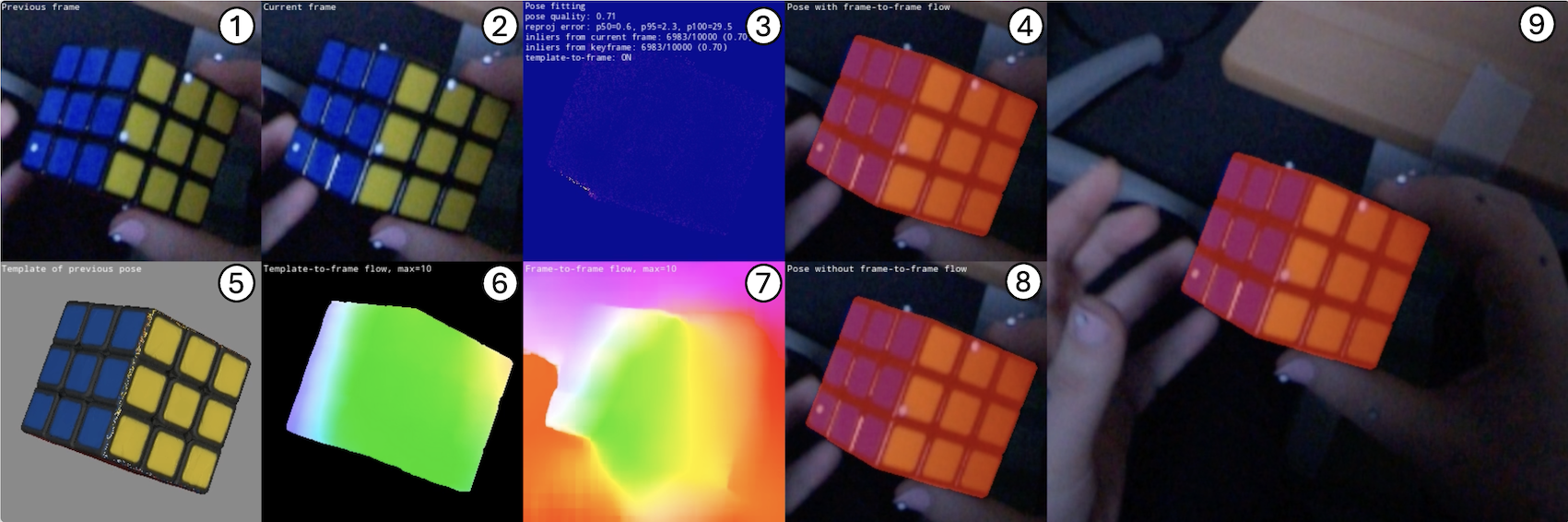}
   \caption{{\bf Visualization of tracking.}
   Tiles \raisebox{-2pt}{\includegraphics[height=4mm]{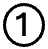}} and \raisebox{-2pt}{\includegraphics[height=4mm]{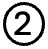}} show crops of the previous and the current frame, respectively.
   Tile \raisebox{-2pt}{\includegraphics[height=4mm]{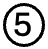}} shows a template rendered with the previous pose estimate.
   Tile \raisebox{-2pt}{\includegraphics[height=4mm]{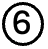}} shows the masked template-to-frame flow, which is visualized for all frames but actually used only when the inlier ratio is below a threshold $\tau_i$.
   Tile \raisebox{-2pt}{\includegraphics[height=4mm]{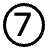}} shows the frame-to-frame flow which is calculated and used in every frame.
   Tile \raisebox{-2pt}{\includegraphics[height=4mm]{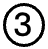}} shows a map of re-projection errors and several lines of text with information about the pose fitting. The second line of text shows the pose quality $q$. The third line shows the statistics of the re-projection error (50th, 75th, and 100th percentile). The fourth line shows the number of used correspondences in the current frame (A), the number of inlier correspondences in the current frame (B), and the inlier ratio in the current frame (B/A). The fifth line shows the number of inlier correspondences in the current frame (B), the number of inlier correspondences in the last keyframe (C), \ie, the frame where we ran the template-to-frame registration last time, and the inlier ratio (B/C) -- the template-to-frame registration is triggered if this inlier ratio is below $\tau_i$. The last line shows whether the template-to-frame registration is triggered.
   Tile~\raisebox{-2pt}{\includegraphics[height=4mm]{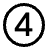}} shows a pose estimated using the full tracking approach described in Sec. 3.3 (\ie, with frame-to-frame flow).
   Tile~\raisebox{-2pt}{\includegraphics[height=4mm]{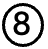}} shows a pose that we would obtain if we ran the template-to-frame registration at every frame without frame-to-frame flow.
   Tile~\raisebox{-2pt}{\includegraphics[height=4mm]{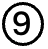}} shows the pose estimate from \raisebox{-2pt}{\includegraphics[height=4mm]{figures/qualitative_supp/label4.png}} in the raw input frame.}
   \label{fig:explaining_visualization}
\end{figure*}

\begin{table}[!t]
\setlength{\tabcolsep}{0.75pt}
\setlength{\tabcolsep}{0.5pt}
\scriptsize
\begin{center}
\begin{tabularx}{1.0\linewidth}{Y YY YY}

\toprule

 & \multicolumn{2}{c}{HOT3D-Aria} & \multicolumn{2}{c}{HOT3D-Quest3} \\

\midrule

 & { ADD} & { ADD-S}  & { ADD} & { ADD-S} \\

\midrule

\multicolumn{1}{l}{dumbbell\_5lb} & 88.1 & 93.5  & 82.3 & 89.5  \\
\multicolumn{1}{l}{cellphone} & 84.8 & 90.0  & 81.2 & 85.9  \\
\multicolumn{1}{l}{mouse} & 88.5 & 93.9  & 86.1 & 91.6  \\
\multicolumn{1}{l}{keyboard} & 89.2 & 92.1  & 89.3 & 91.8  \\
\multicolumn{1}{l}{mug\_white} & 89.0 & 94.6  & 84.5 & 90.9  \\
\multicolumn{1}{l}{mug\_patterned} & 86.0 & 92.6  & 88.1 & 94.1  \\
\multicolumn{1}{l}{mug\_patterned} &  87.9 & 93.4  & 88.9 & 94.0  \\
\multicolumn{1}{l}{bowl} & 62.6 & 90.5  & 64.9 & 88.5 \\
\multicolumn{1}{l}{plate\_bamboo} & 62.7 & 80.0  & 63.7 & 77.8 \\
\multicolumn{1}{l}{flask} & 85.4 & 92.5  & 85.0 & 91.7  \\
\multicolumn{1}{l}{vase} & 85.5 & 92.4  & 87.8 & 93.4  \\
\multicolumn{1}{l}{spoon\_wooden} & 70.4 & 77.3  & 58.4 & 65.5  \\
\multicolumn{1}{l}{spatula\_red} & 65.7 & 72.3 & 57.6 & 66.1  \\
\multicolumn{1}{l}{potato\_masher} & 85.4 & 91.4  & 78.8 & 88.0  \\
\multicolumn{1}{l}{holder\_black} & 84.4 & 92.6  & 77.4 & 87.8  \\
\multicolumn{1}{l}{holder\_gray} & 88.7 & 93.7  & 85.9 & 92.7  \\
\multicolumn{1}{l}{birdhouse\_toy} & 88.4 & 93.8  & 91.2 & 94.4  \\
\multicolumn{1}{l}{dino\_toy} & 87.1 & 92.9  & 85.1 & 92.2  \\
\multicolumn{1}{l}{dvd\_remote} & 74.4 & 79.6  & 78.6 & 85.1  \\
\multicolumn{1}{l}{whiteboard\_eraser} & 85.6 & 90.8  & 84.4 & 90.1  \\
\multicolumn{1}{l}{whiteboard\_marker} & 85.2 & 91.5  & 83.8 & 89.1  \\
\multicolumn{1}{l}{carton\_milk} & 89.4 & 93.4  & 85.9 & 92.8  \\
\multicolumn{1}{l}{carton\_oj} & 93.6 & 95.9  & 86.2 & 93.1  \\
\multicolumn{1}{l}{bottle\_mustard} & 88.4 & 94.1  & 79.5 & 89.4  \\
\multicolumn{1}{l}{bottle\_ranch} & 86.8 & 92.8  & 80.9 & 90.3  \\
\multicolumn{1}{l}{bottle\_bbq} & 76.8 & 88.5  & 81.2 & 90.9  \\
\multicolumn{1}{l}{can\_soup} & 89.6 & 94.3  & 84.5 & 90.9  \\
\multicolumn{1}{l}{can\_parmesan} & 91.0 & 94.6  & 88.9 & 93.8  \\
\multicolumn{1}{l}{can\_tomato\_sauce} & 90.6 & 94.8  & 85.8 & 92.3  \\
\multicolumn{1}{l}{food\_waffles} & 85.1 & 92.2  & 86.3 & 91.6  \\
\multicolumn{1}{l}{food\_vegetables} & 87.5 & 92.6  & 85.9 & 91.3  \\
\multicolumn{1}{l}{puzzle\_toy} & 94.2 & 96.6  & 91.8 & 95.3  \\
\multicolumn{1}{l}{aria\_small} & 65.7 & 81.1  & 70.9 & 84.9  \\
\midrule

\multicolumn{1}{l}{Average} & 83.7 & 90.7  & 81.5 & 89.0  \\

\bottomrule
 
\end{tabularx}
\end{center}
\caption{\textbf{Tracking performance on the HOT3D dataset~\cite{hot3d}.} The accuracy is measured by AUC \wrt the ADD and ADD-S pose error functions for multiple
thresholds ranging from 0.1 cm to 10 cm (as in Table 2 of the main paper).
}\label{tab:hot3d}
\end{table}

\section{Qualitative refinement results} \label{app:ref_qual}
Fig.~\ref{fig:lmo}--\ref{fig:itodd} show multiple object pose refinement results (using only one iteration) on the seven core BOP datasets~\cite{hodan2023bop}.
Each row presents one sample. The input image is shown in the first column, and the coarse pose from the fast template retrieval of FoundPose~\cite{foundPose} is shown in the second column. The third and fourth columns show the predictions of our network (a mask of visible template pixels and an optical flow field), which can be used to remap pixels from the template to the input image as shown in the fifth column. The last column presents the initial pose (in blue), and the final pose estimated by P\textit{n}P-RANSAC from 2D-3D correspondences which are constructed from the network predictions (in red). As can be seen in the second and third columns in all the figures, our method is robust to occlusions and accurately estimates the visible mask, which results in an accurate pose estimate. Furthermore, many of the samples (\eg, the first two rows of Fig.~\ref{fig:tless}) also demonstrate that our method can correct large pose deviations.

\section{Qualitative tracking results} \label{app:track_qual}
\begin{itemize}
    \item Tracking results on the YCB-V dataset~\cite{xiang2017posecnn}:\\
    \texttt{https://youtu.be/6B1ZOQkvxH4}
    \item Tracking results on the HOT3D-Aria dataset~\cite{hot3d}:\\\texttt{https://youtu.be/74NXe03ySdM};
    \item Tracking results on the HOT3D-Quest3 dataset~\cite{hot3d}:\\\texttt{https://youtu.be/bjQWS9vgPT0}.
\end{itemize}

Each video shows the inputs, the model-to-frame\footnote{The terms ``model-to-frame registration'' and ``template-to-frame registration'' are used interchangeably.} flow, the frame-to-frame flow, the pose fitting statistics, and the estimated poses for sample sequences (see Fig.~\ref{fig:explaining_visualization} for details). The poses are initialized using the template retrieval approach from FoundPose~\cite{foundPose}.

As detailed in Sec.~\ref{sec:tracking_pipeline}, the model-to-frame registration is triggered when the inlier ratio is below $\tau_i=0.8$ (\ie, when less than 80\% of the inlier correspondences from the last model-to-frame registration are still inliers in the current frame).
The frequency of triggering the model-to-frame registration depends on dynamics of the scene -- if the object pose and the viewpoint change slowly, as in YCB-V~\cite{xiang2017posecnn}, the model-to-frame registration is triggered less often than in the more dynamic sequences from HOT3D~\cite{hot3d}, particularly when objects are manipulated by hands. The frequency can be reduced by lowering $\tau_i$ for the price of potentially introducing a drift from the actual pose. In every frame, we use the maximum of 10K correspondences (if more are available, we sample a random subset).

\setlength\plotwidth{2.5cm}
\setlength\lineskip{1.5pt}
\setlength\tabcolsep{1.5pt} 

\begin{figure*}[!t]
\begin{center}
{\small
\begin{tabular}{
>{\centering\arraybackslash}m{\plotwidth}%
>{\centering\arraybackslash}m{\plotwidth}%
>{\centering\arraybackslash}m{\plotwidth}
>{\centering\arraybackslash}m{\plotwidth}%
>{\centering\arraybackslash}m{\plotwidth}%
>{\centering\arraybackslash}m{\plotwidth}
}

Input image & Template  & Predicted mask & Predicted flow & Warped template  & Estimated pose \\ 
\frame{\includegraphics[width=\plotwidth,]{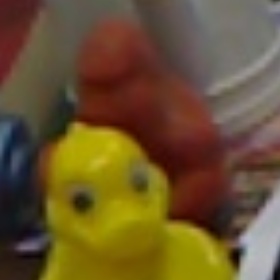}}&
\frame{\includegraphics[width=\plotwidth, ]{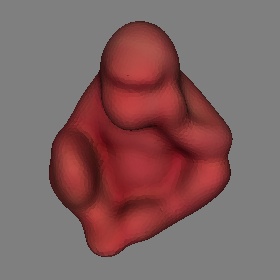}} &
\frame{\includegraphics[width=\plotwidth, ]{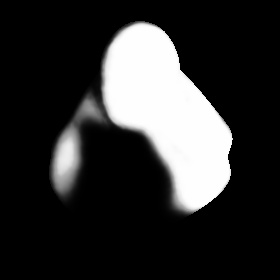}} &
\frame{\includegraphics[width=\plotwidth,]{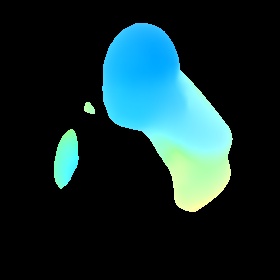}} &
\frame{\includegraphics[width=\plotwidth, ]{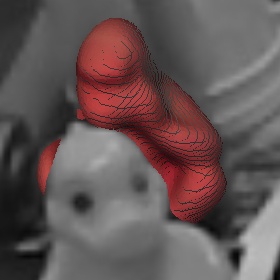}} &
\frame{\includegraphics[width=\plotwidth,]{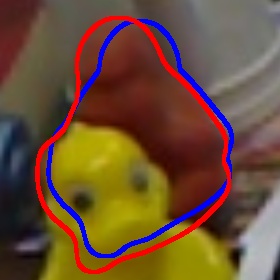}} \\

\frame{\includegraphics[width=\plotwidth,]{figures/qualitative_supp/lmo/global_batch_id821367_batch_id02_sample3_query.jpg}}&
\frame{\includegraphics[width=\plotwidth, ]{figures/qualitative_supp/lmo/global_batch_id821367_batch_id02_sample3_template.jpg}} &
\frame{\includegraphics[width=\plotwidth, ]{figures/qualitative_supp/lmo/global_batch_id821367_batch_id02_sample3_pred_mask.jpg}} &
\frame{\includegraphics[width=\plotwidth,]{figures/qualitative_supp/lmo/global_batch_id821367_batch_id02_sample3_flow.jpg}} &
\frame{\includegraphics[width=\plotwidth, ]{figures/qualitative_supp/lmo/global_batch_id821367_batch_id02_sample3_warped_template.jpg}} &
\frame{\includegraphics[width=\plotwidth,]{figures/qualitative_supp/lmo/global_batch_id821367_batch_id02_sample3_prediction.jpg}} \\

\frame{\includegraphics[width=\plotwidth,]{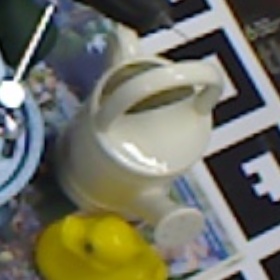}}&
\frame{\includegraphics[width=\plotwidth, ]{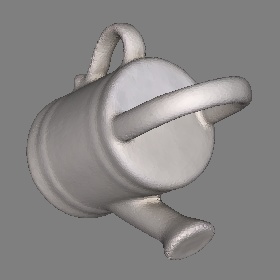}} &
\frame{\includegraphics[width=\plotwidth, ]{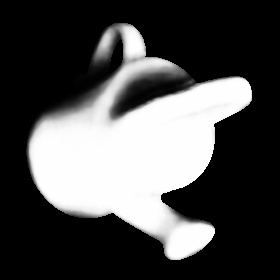}} &
\frame{\includegraphics[width=\plotwidth,]{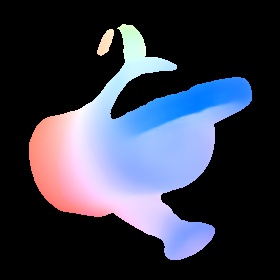}} &
\frame{\includegraphics[width=\plotwidth, ]{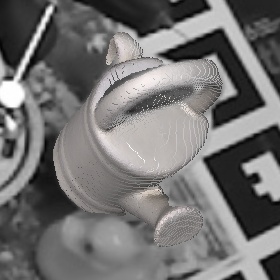}} &
\frame{\includegraphics[width=\plotwidth,]{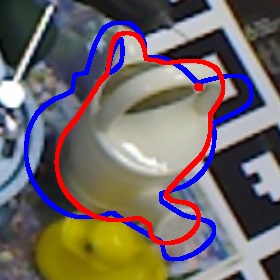}} \\

\frame{\includegraphics[width=\plotwidth,]{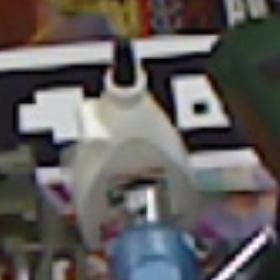}}&
\frame{\includegraphics[width=\plotwidth, ]{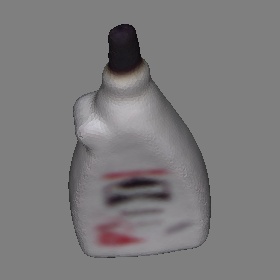}} &
\frame{\includegraphics[width=\plotwidth, ]{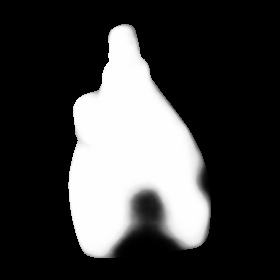}} &
\frame{\includegraphics[width=\plotwidth,]{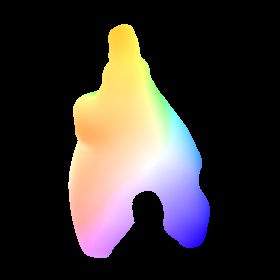}} &
\frame{\includegraphics[width=\plotwidth, ]{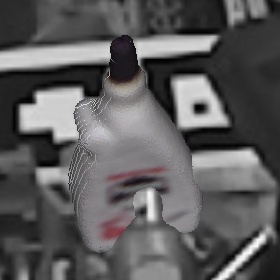}} &
\frame{\includegraphics[width=\plotwidth,]{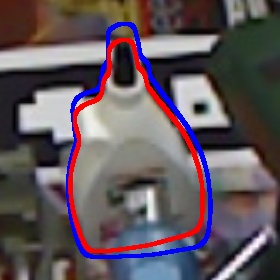}} \\

\frame{\includegraphics[width=\plotwidth,]{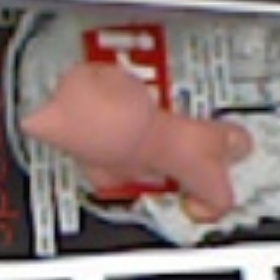}}&
\frame{\includegraphics[width=\plotwidth, ]{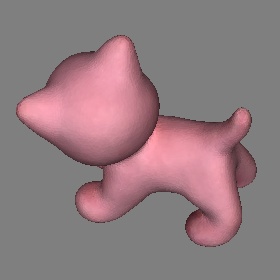}} &
\frame{\includegraphics[width=\plotwidth, ]{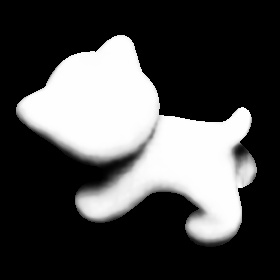}} &
\frame{\includegraphics[width=\plotwidth,]{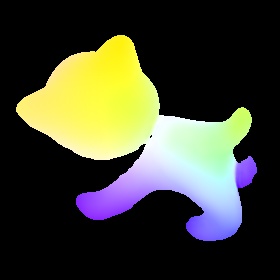}} &
\frame{\includegraphics[width=\plotwidth, ]{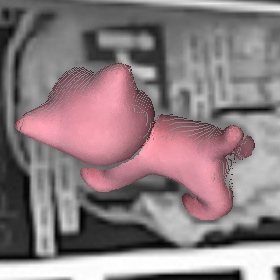}} &
\frame{\includegraphics[width=\plotwidth,]{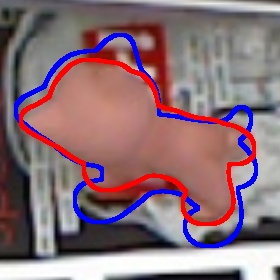}} \\

\frame{\includegraphics[width=\plotwidth,]{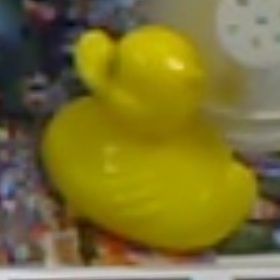}}&
\frame{\includegraphics[width=\plotwidth, ]{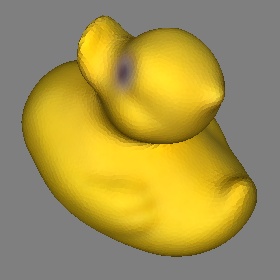}} &
\frame{\includegraphics[width=\plotwidth, ]{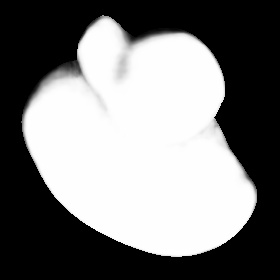}} &
\frame{\includegraphics[width=\plotwidth,]{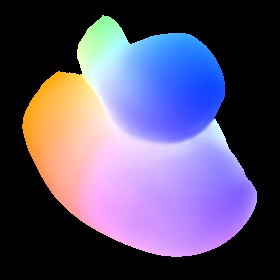}} &
\frame{\includegraphics[width=\plotwidth, ]{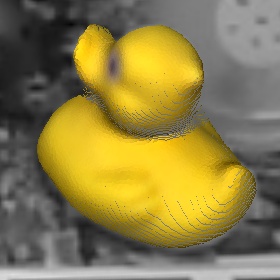}} &
\frame{\includegraphics[width=\plotwidth,]{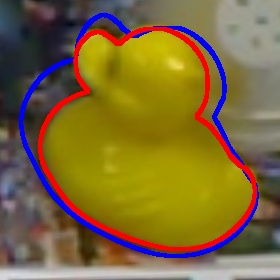}} \\

\frame{\includegraphics[width=\plotwidth,]{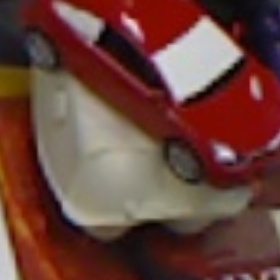}}&
\frame{\includegraphics[width=\plotwidth, ]{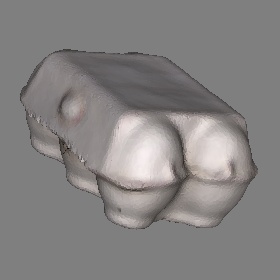}} &
\frame{\includegraphics[width=\plotwidth, ]{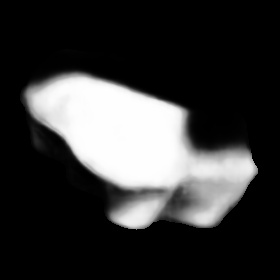}} &
\frame{\includegraphics[width=\plotwidth,]{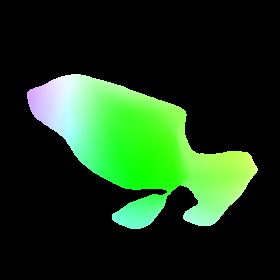}} &
\frame{\includegraphics[width=\plotwidth, ]{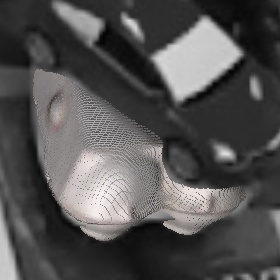}} &
\frame{\includegraphics[width=\plotwidth,]{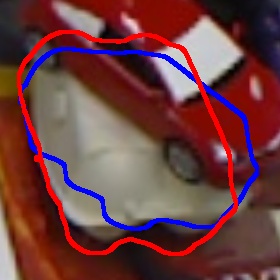}} \\

\frame{\includegraphics[width=\plotwidth,]{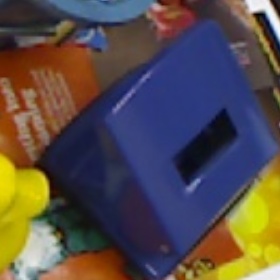}}&
\frame{\includegraphics[width=\plotwidth, ]{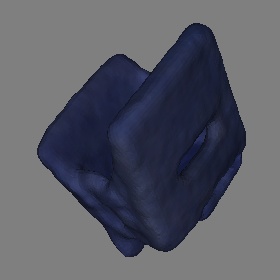}} &
\frame{\includegraphics[width=\plotwidth, ]{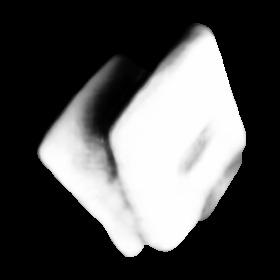}} &
\frame{\includegraphics[width=\plotwidth,]{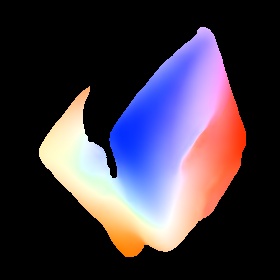}} &
\frame{\includegraphics[width=\plotwidth, ]{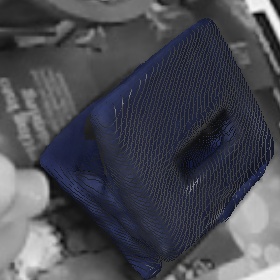}} &
\frame{\includegraphics[width=\plotwidth,]{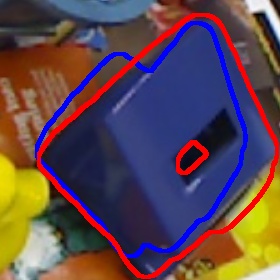}} \\

\end{tabular}
}

    \captionof{figure}{ 
     \textbf{Example results of the GoTrack refiner on LM-O~\cite{brachmann-eccv14-learning6dobjectposeestimation}}.
     }
    \label{fig:lmo}
\end{center}

\end{figure*}

\setlength\plotwidth{2.5cm}
\setlength\lineskip{1.5pt}
\setlength\tabcolsep{1.5pt} 

\begin{figure*}[!t]
\begin{center}
{\small
\begin{tabular}{
>{\centering\arraybackslash}m{\plotwidth}%
>{\centering\arraybackslash}m{\plotwidth}%
>{\centering\arraybackslash}m{\plotwidth}
>{\centering\arraybackslash}m{\plotwidth}%
>{\centering\arraybackslash}m{\plotwidth}%
>{\centering\arraybackslash}m{\plotwidth}
}

Input image & Template  & Predicted mask & Predicted flow & Warped template & Estimated pose \\

\frame{\includegraphics[width=\plotwidth,]{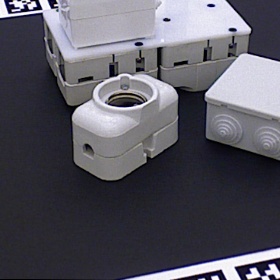}}&
\frame{\includegraphics[width=\plotwidth, ]{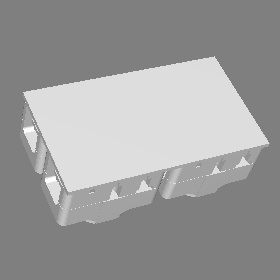}} &
\frame{\includegraphics[width=\plotwidth, ]{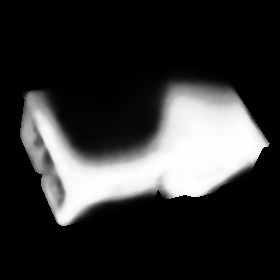}} &
\frame{\includegraphics[width=\plotwidth,]{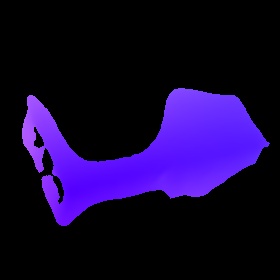}} &
\frame{\includegraphics[width=\plotwidth, ]{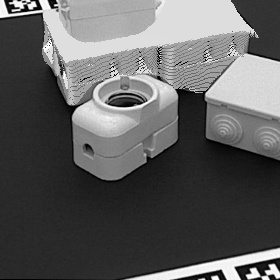}} &
\frame{\includegraphics[width=\plotwidth,]{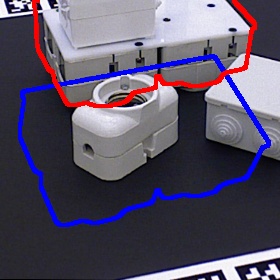}} \\

\frame{\includegraphics[width=\plotwidth,]{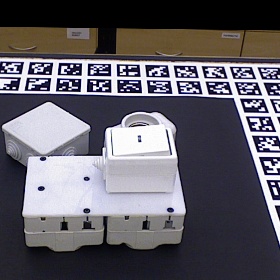}}&
\frame{\includegraphics[width=\plotwidth, ]{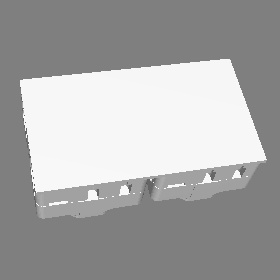}} &
\frame{\includegraphics[width=\plotwidth, ]{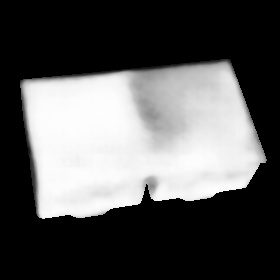}} &
\frame{\includegraphics[width=\plotwidth,]{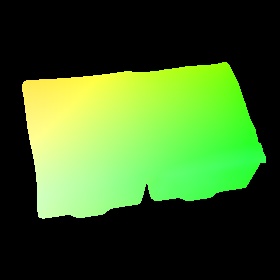}} &
\frame{\includegraphics[width=\plotwidth, ]{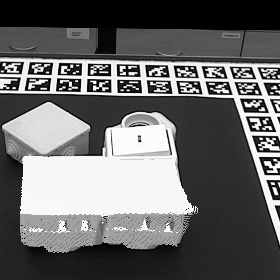}} &
\frame{\includegraphics[width=\plotwidth,]{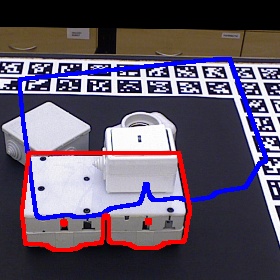}} \\

\frame{\includegraphics[width=\plotwidth,]{figures/qualitative_supp/tless/global_batch_id821367_batch_id289query.jpg}}&
\frame{\includegraphics[width=\plotwidth, ]{figures/qualitative_supp/tless/global_batch_id821367_batch_id289template.jpg}} &
\frame{\includegraphics[width=\plotwidth, ]{figures/qualitative_supp/tless/global_batch_id821367_batch_id289pred_mask.jpg}} &
\frame{\includegraphics[width=\plotwidth,]{figures/qualitative_supp/tless/global_batch_id821367_batch_id289flow.jpg}} &
\frame{\includegraphics[width=\plotwidth, ]{figures/qualitative_supp/tless/global_batch_id821367_batch_id289warped_template.jpg}} &
\frame{\includegraphics[width=\plotwidth,]{figures/qualitative_supp/tless/global_batch_id821367_batch_id289prediction.jpg}} \\

\frame{\includegraphics[width=\plotwidth,]{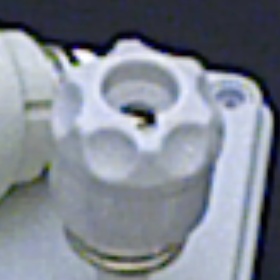}}&
\frame{\includegraphics[width=\plotwidth, ]{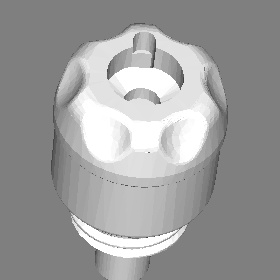}} &
\frame{\includegraphics[width=\plotwidth, ]{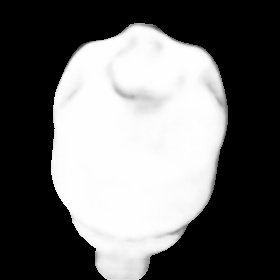}} &
\frame{\includegraphics[width=\plotwidth,]{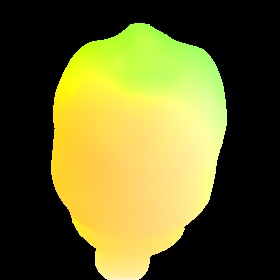}} &
\frame{\includegraphics[width=\plotwidth, ]{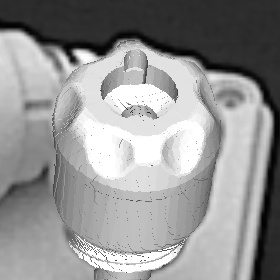}} &
\frame{\includegraphics[width=\plotwidth,]{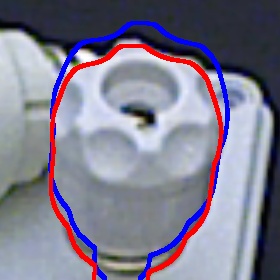}} \\

\frame{\includegraphics[width=\plotwidth,]{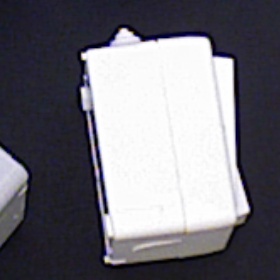}}&
\frame{\includegraphics[width=\plotwidth, ]{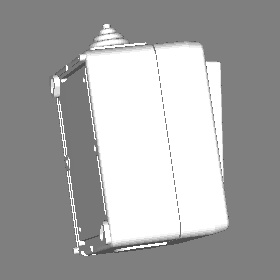}} &
\frame{\includegraphics[width=\plotwidth, ]{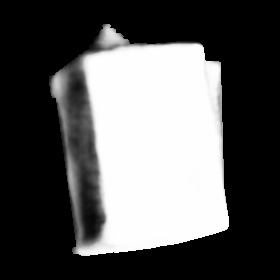}} &
\frame{\includegraphics[width=\plotwidth,]{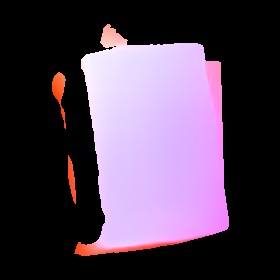}} &
\frame{\includegraphics[width=\plotwidth, ]{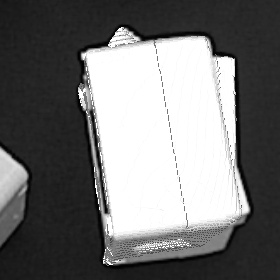}} &
\frame{\includegraphics[width=\plotwidth,]{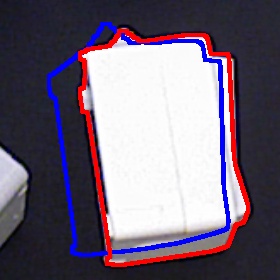}} \\

\frame{\includegraphics[width=\plotwidth,]{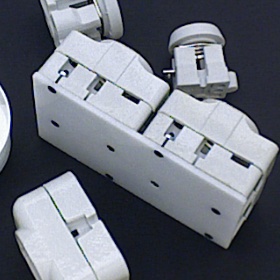}}&
\frame{\includegraphics[width=\plotwidth, ]{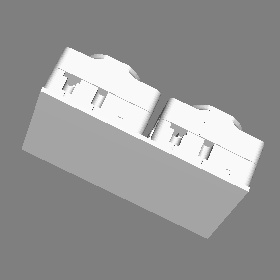}} &
\frame{\includegraphics[width=\plotwidth, ]{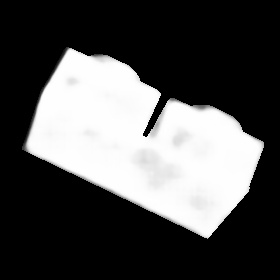}} &
\frame{\includegraphics[width=\plotwidth,]{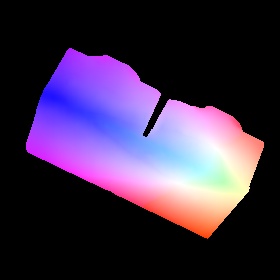}} &
\frame{\includegraphics[width=\plotwidth, ]{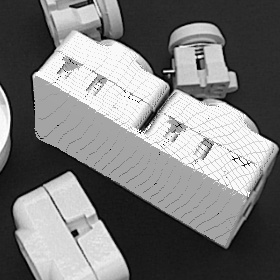}} &
\frame{\includegraphics[width=\plotwidth,]{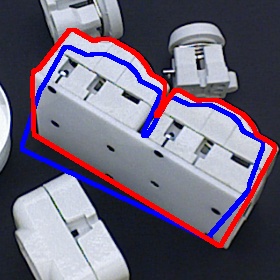}} \\

\frame{\includegraphics[width=\plotwidth,]{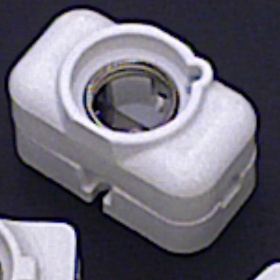}}&
\frame{\includegraphics[width=\plotwidth, ]{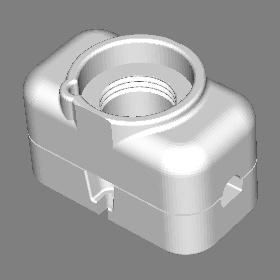}} &
\frame{\includegraphics[width=\plotwidth, ]{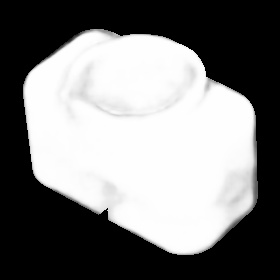}} &
\frame{\includegraphics[width=\plotwidth,]{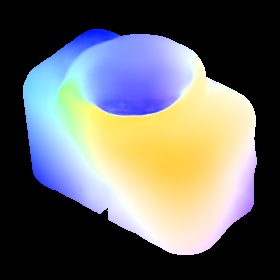}} &
\frame{\includegraphics[width=\plotwidth, ]{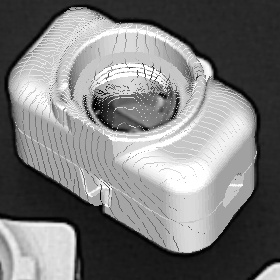}} &
\frame{\includegraphics[width=\plotwidth,]{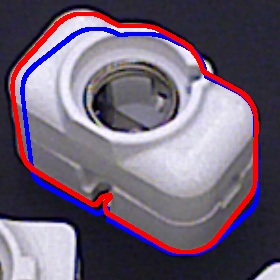}} \\

\frame{\includegraphics[width=\plotwidth,]{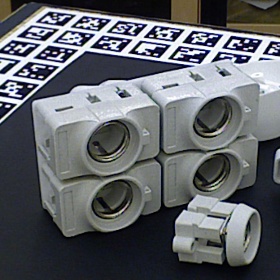}}&
\frame{\includegraphics[width=\plotwidth, ]{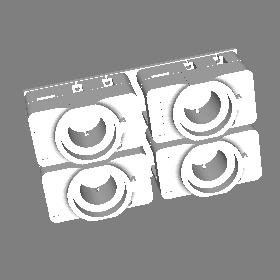}} &
\frame{\includegraphics[width=\plotwidth, ]{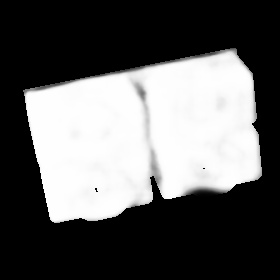}} &
\frame{\includegraphics[width=\plotwidth,]{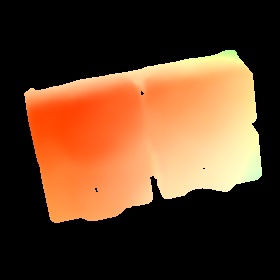}} &
\frame{\includegraphics[width=\plotwidth, ]{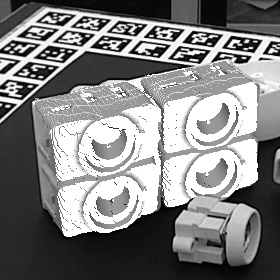}} &
\frame{\includegraphics[width=\plotwidth,]{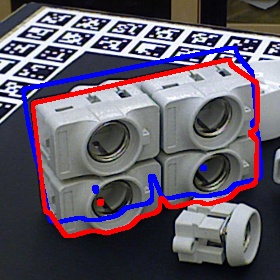}} \\

\end{tabular}
}

    \captionof{figure}{ 
     \textbf{Example results of the GoTrack refiner on T-LESS~\cite{hodan-wacv17-tless}}.
     }
    \label{fig:tless}
\end{center}
\end{figure*}

\setlength\plotwidth{2.5cm}
\setlength\lineskip{1.5pt}
\setlength\tabcolsep{1.5pt} 

\begin{figure*}[!t]
\begin{center}
{\small
\begin{tabular}{
>{\centering\arraybackslash}m{\plotwidth}%
>{\centering\arraybackslash}m{\plotwidth}%
>{\centering\arraybackslash}m{\plotwidth}
>{\centering\arraybackslash}m{\plotwidth}%
>{\centering\arraybackslash}m{\plotwidth}%
>{\centering\arraybackslash}m{\plotwidth}
}

Input image & Template  & Predicted mask & Predicted flow & Warped template & Estimated pose \\ 
\frame{\includegraphics[width=\plotwidth,]{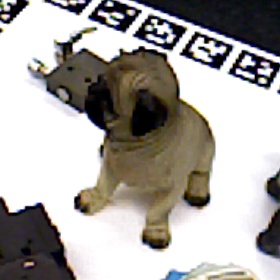}}&
\frame{\includegraphics[width=\plotwidth, ]{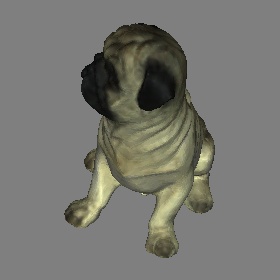}} &
\frame{\includegraphics[width=\plotwidth, ]{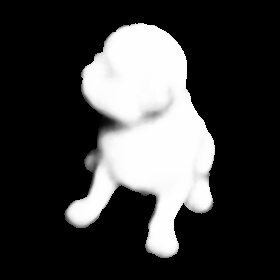}} &
\frame{\includegraphics[width=\plotwidth,]{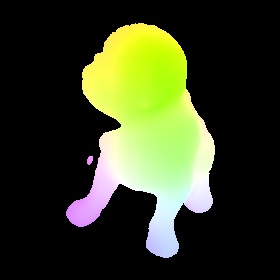}} &
\frame{\includegraphics[width=\plotwidth, ]{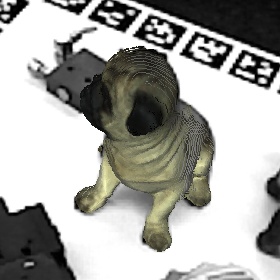}} &
\frame{\includegraphics[width=\plotwidth,]{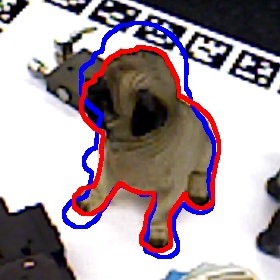}} \\

\frame{\includegraphics[width=\plotwidth,]{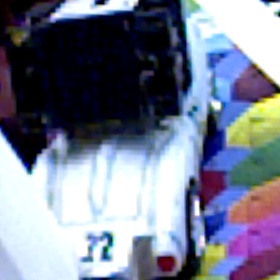}}&
\frame{\includegraphics[width=\plotwidth, ]{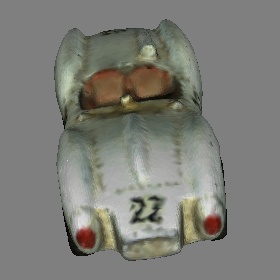}} &
\frame{\includegraphics[width=\plotwidth, ]{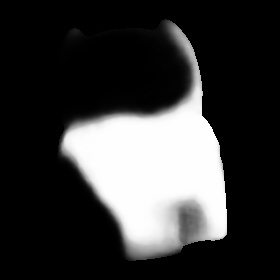}} &
\frame{\includegraphics[width=\plotwidth,]{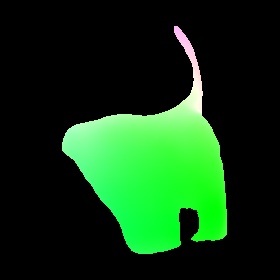}} &
\frame{\includegraphics[width=\plotwidth, ]{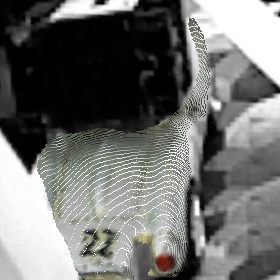}} &
\frame{\includegraphics[width=\plotwidth,]{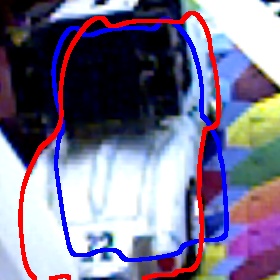}} \\

\frame{\includegraphics[width=\plotwidth,]{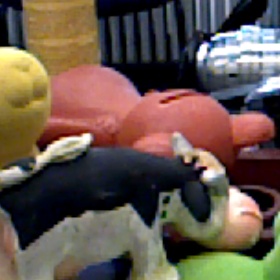}}&
\frame{\includegraphics[width=\plotwidth, ]{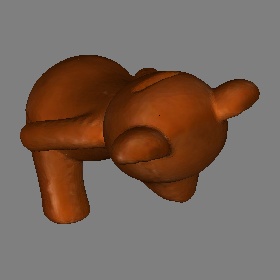}} &
\frame{\includegraphics[width=\plotwidth, ]{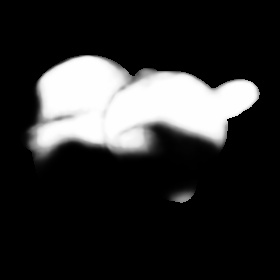}} &
\frame{\includegraphics[width=\plotwidth,]{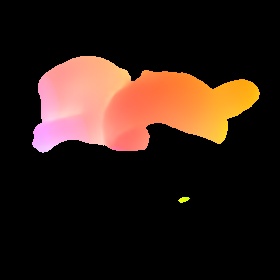}} &
\frame{\includegraphics[width=\plotwidth, ]{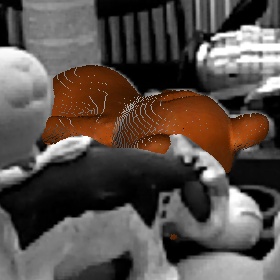}} &
\frame{\includegraphics[width=\plotwidth,]{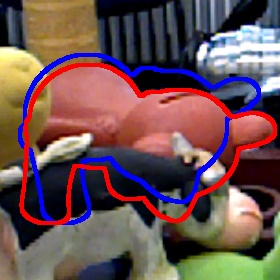}} \\

\frame{\includegraphics[width=\plotwidth,]{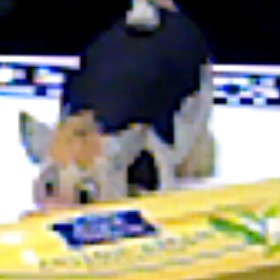}}&
\frame{\includegraphics[width=\plotwidth, ]{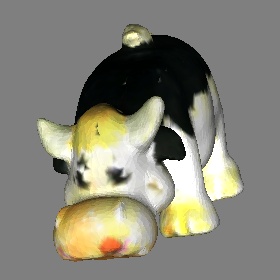}} &
\frame{\includegraphics[width=\plotwidth, ]{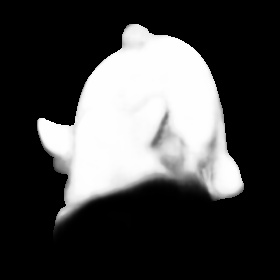}} &
\frame{\includegraphics[width=\plotwidth,]{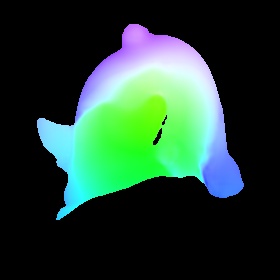}} &
\frame{\includegraphics[width=\plotwidth, ]{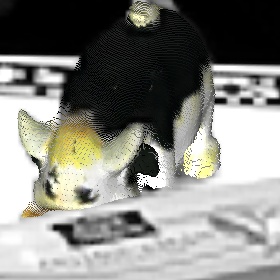}} &
\frame{\includegraphics[width=\plotwidth,]{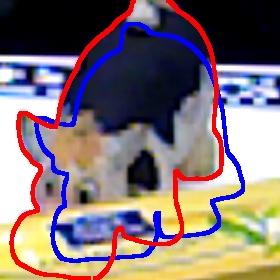}} \\

\frame{\includegraphics[width=\plotwidth,]{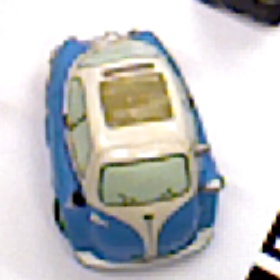}}&
\frame{\includegraphics[width=\plotwidth, ]{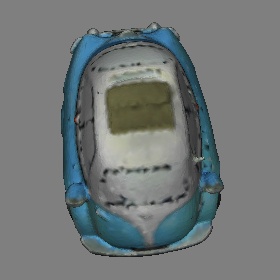}} &
\frame{\includegraphics[width=\plotwidth, ]{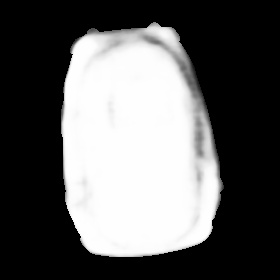}} &
\frame{\includegraphics[width=\plotwidth,]{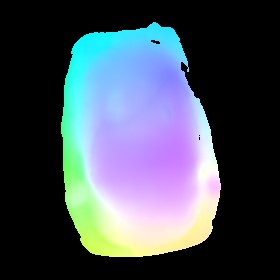}} &
\frame{\includegraphics[width=\plotwidth, ]{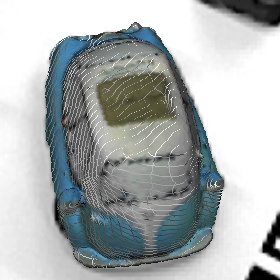}} &
\frame{\includegraphics[width=\plotwidth,]{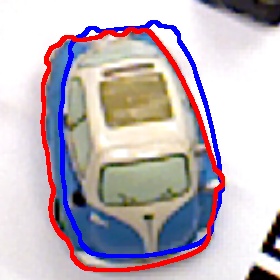}} \\

\frame{\includegraphics[width=\plotwidth,]{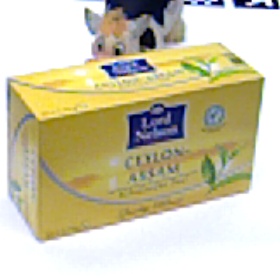}}&
\frame{\includegraphics[width=\plotwidth, ]{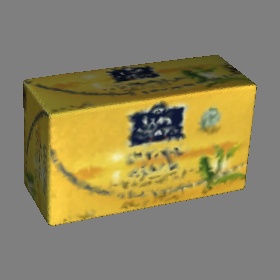}} &
\frame{\includegraphics[width=\plotwidth, ]{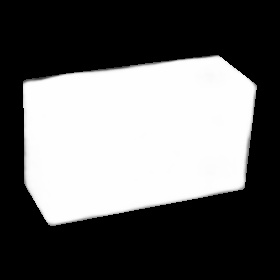}} &
\frame{\includegraphics[width=\plotwidth,]{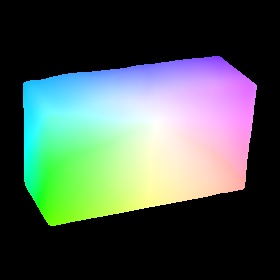}} &
\frame{\includegraphics[width=\plotwidth, ]{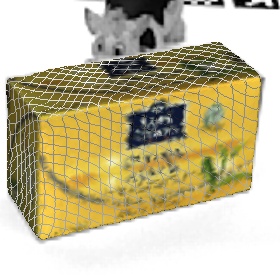}} &
\frame{\includegraphics[width=\plotwidth,]{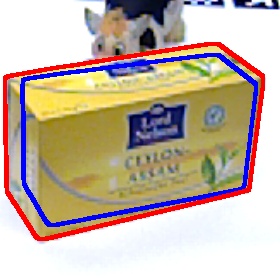}} \\

\frame{\includegraphics[width=\plotwidth,]{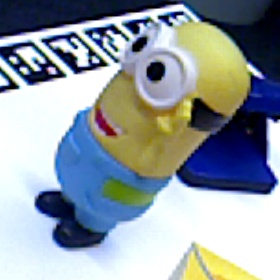}}&
\frame{\includegraphics[width=\plotwidth, ]{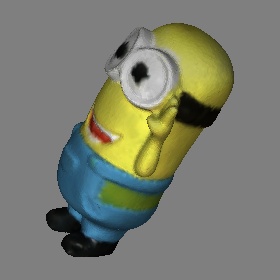}} &
\frame{\includegraphics[width=\plotwidth, ]{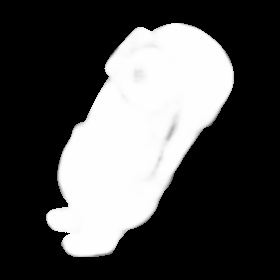}} &
\frame{\includegraphics[width=\plotwidth,]{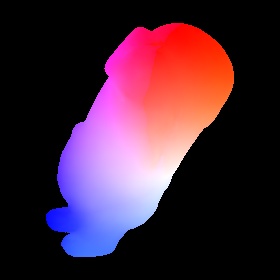}} &
\frame{\includegraphics[width=\plotwidth, ]{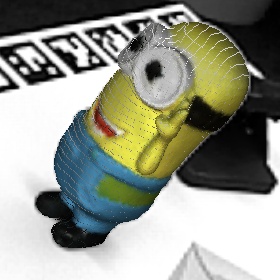}} &
\frame{\includegraphics[width=\plotwidth,]{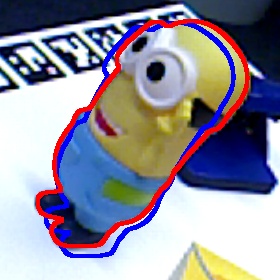}} \\

\frame{\includegraphics[width=\plotwidth,]{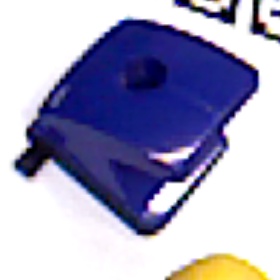}}&
\frame{\includegraphics[width=\plotwidth, ]{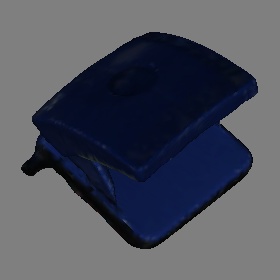}} &
\frame{\includegraphics[width=\plotwidth, ]{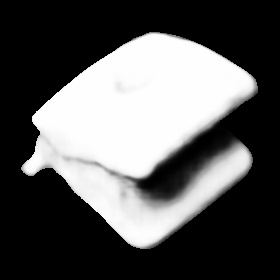}} &
\frame{\includegraphics[width=\plotwidth,]{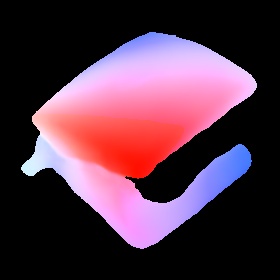}} &
\frame{\includegraphics[width=\plotwidth, ]{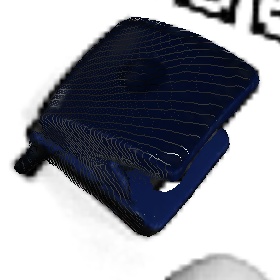}} &
\frame{\includegraphics[width=\plotwidth,]{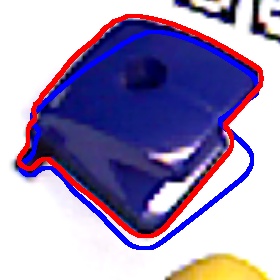}} \\

\end{tabular}
}

    \captionof{figure}{ 
     \textbf{Example results of the GoTrack refiner on HB~\cite{kaskman2019homebreweddb}}.
     }
    \label{fig:hb}
\end{center}
\end{figure*}

\setlength\plotwidth{2.5cm}
\setlength\lineskip{1.5pt}
\setlength\tabcolsep{1.5pt} 

\begin{figure*}[!t]
\begin{center}
{\small
\begin{tabular}{
>{\centering\arraybackslash}m{\plotwidth}%
>{\centering\arraybackslash}m{\plotwidth}%
>{\centering\arraybackslash}m{\plotwidth}
>{\centering\arraybackslash}m{\plotwidth}%
>{\centering\arraybackslash}m{\plotwidth}%
>{\centering\arraybackslash}m{\plotwidth}
}

Input image & Template  & Predicted mask & Predicted flow & Warped template & Estimated pose \\ 

\frame{\includegraphics[width=\plotwidth,]{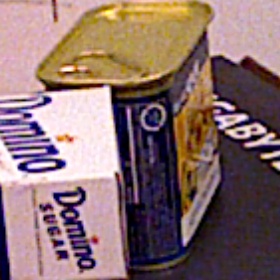}}&
\frame{\includegraphics[width=\plotwidth, ]{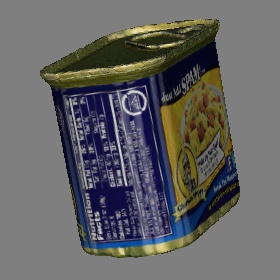}} &
\frame{\includegraphics[width=\plotwidth, ]{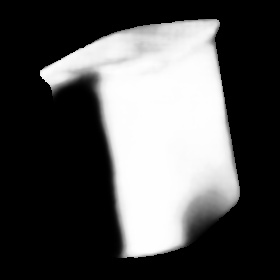}} &
\frame{\includegraphics[width=\plotwidth,]{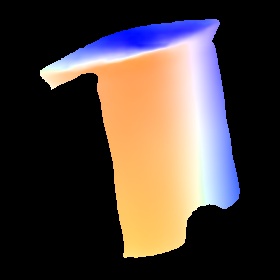}} &
\frame{\includegraphics[width=\plotwidth, ]{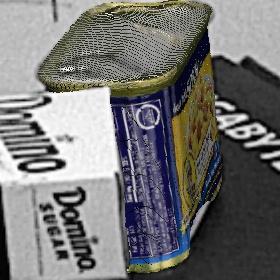}} &
\frame{\includegraphics[width=\plotwidth,]{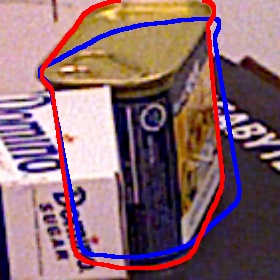}} \\

\frame{\includegraphics[width=\plotwidth,]{figures/qualitative_supp/ycbv/global_batch_id821367_batch_id158_sample0_query.jpg}}&
\frame{\includegraphics[width=\plotwidth, ]{figures/qualitative_supp/ycbv/global_batch_id821367_batch_id158_sample0_template.jpg}} &
\frame{\includegraphics[width=\plotwidth, ]{figures/qualitative_supp/ycbv/global_batch_id821367_batch_id158_sample0_pred_mask.jpg}} &
\frame{\includegraphics[width=\plotwidth,]{figures/qualitative_supp/ycbv/global_batch_id821367_batch_id158_sample0_flow.jpg}} &
\frame{\includegraphics[width=\plotwidth, ]{figures/qualitative_supp/ycbv/global_batch_id821367_batch_id158_sample0_warped_template.jpg}} &
\frame{\includegraphics[width=\plotwidth,]{figures/qualitative_supp/ycbv/global_batch_id821367_batch_id158_sample0_prediction.jpg}} \\

\frame{\includegraphics[width=\plotwidth,]{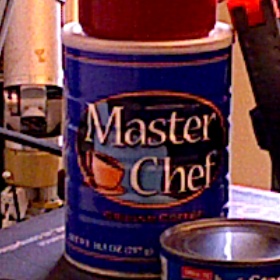}}&
\frame{\includegraphics[width=\plotwidth, ]{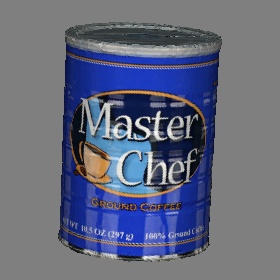}} &
\frame{\includegraphics[width=\plotwidth, ]{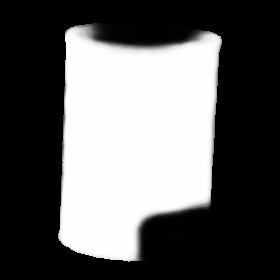}} &
\frame{\includegraphics[width=\plotwidth,]{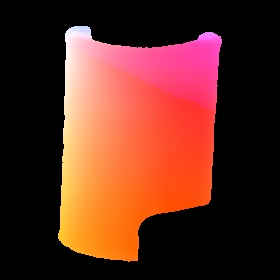}} &
\frame{\includegraphics[width=\plotwidth, ]{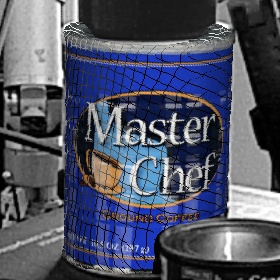}} &
\frame{\includegraphics[width=\plotwidth,]{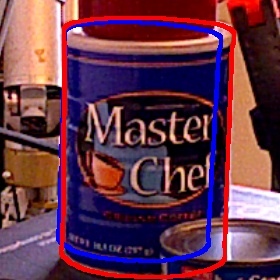}} \\

\frame{\includegraphics[width=\plotwidth,]{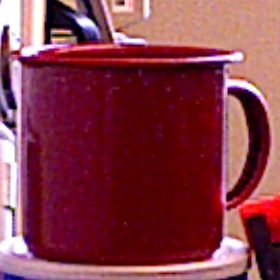}}&
\frame{\includegraphics[width=\plotwidth, ]{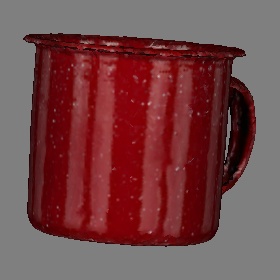}} &
\frame{\includegraphics[width=\plotwidth, ]{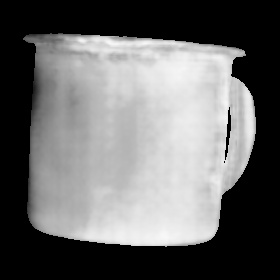}} &
\frame{\includegraphics[width=\plotwidth,]{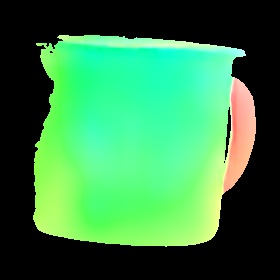}} &
\frame{\includegraphics[width=\plotwidth, ]{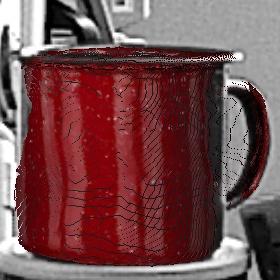}} &
\frame{\includegraphics[width=\plotwidth,]{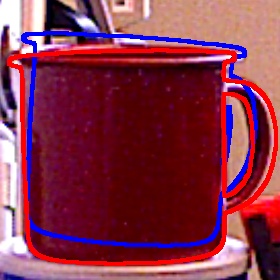}} \\

\frame{\includegraphics[width=\plotwidth,]{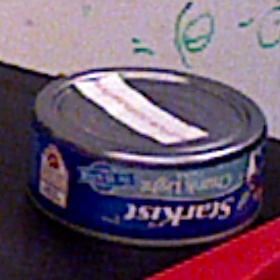}}&
\frame{\includegraphics[width=\plotwidth, ]{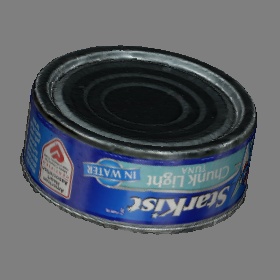}} &
\frame{\includegraphics[width=\plotwidth, ]{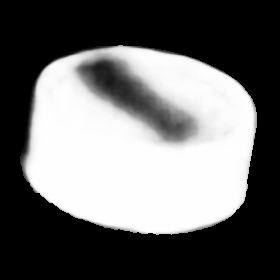}} &
\frame{\includegraphics[width=\plotwidth,]{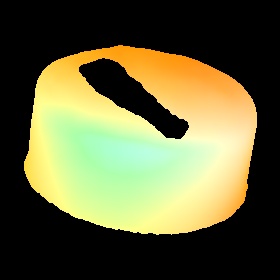}} &
\frame{\includegraphics[width=\plotwidth, ]{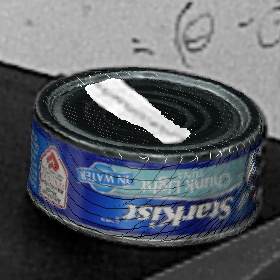}} &
\frame{\includegraphics[width=\plotwidth,]{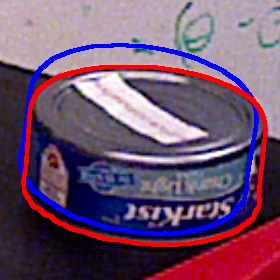}} \\

\frame{\includegraphics[width=\plotwidth,]{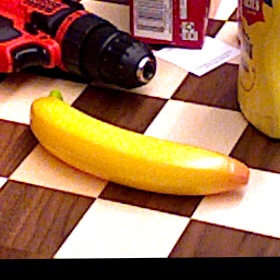}}&
\frame{\includegraphics[width=\plotwidth, ]{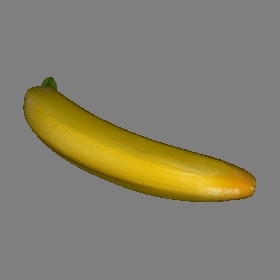}} &
\frame{\includegraphics[width=\plotwidth, ]{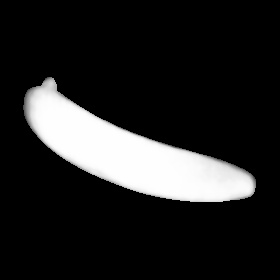}} &
\frame{\includegraphics[width=\plotwidth,]{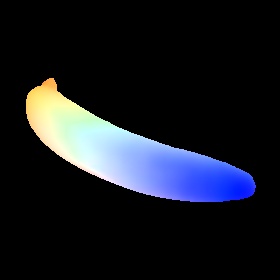}} &
\frame{\includegraphics[width=\plotwidth, ]{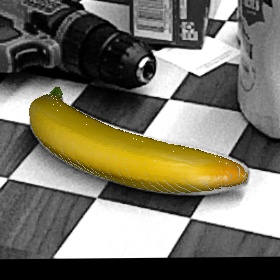}} &
\frame{\includegraphics[width=\plotwidth,]{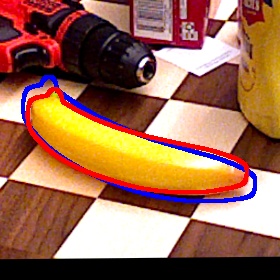}} \\

\frame{\includegraphics[width=\plotwidth,]{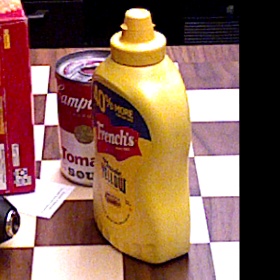}}&
\frame{\includegraphics[width=\plotwidth, ]{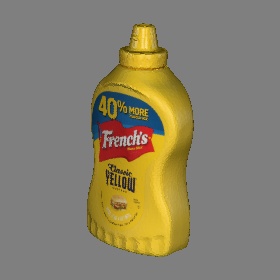}} &
\frame{\includegraphics[width=\plotwidth, ]{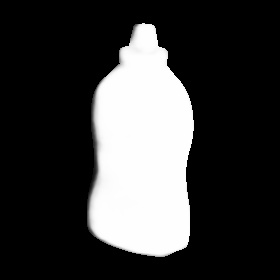}} &
\frame{\includegraphics[width=\plotwidth,]{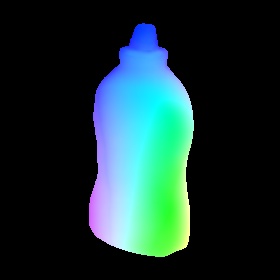}} &
\frame{\includegraphics[width=\plotwidth, ]{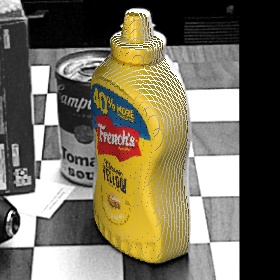}} &
\frame{\includegraphics[width=\plotwidth,]{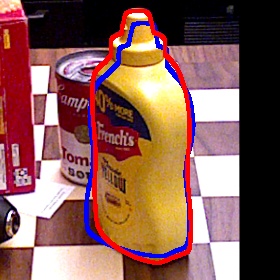}} \\

\frame{\includegraphics[width=\plotwidth,]{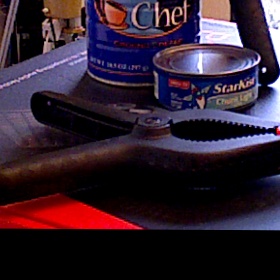}}&
\frame{\includegraphics[width=\plotwidth, ]{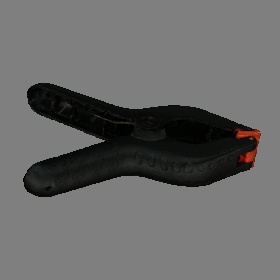}} &
\frame{\includegraphics[width=\plotwidth, ]{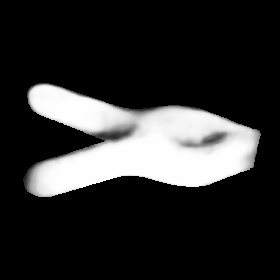}} &
\frame{\includegraphics[width=\plotwidth,]{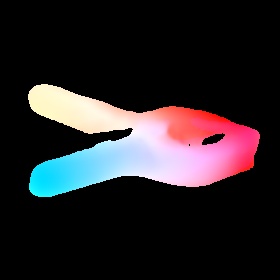}} &
\frame{\includegraphics[width=\plotwidth, ]{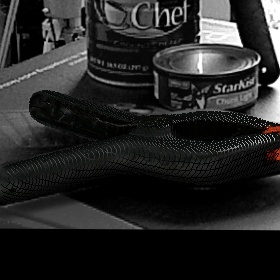}} &
\frame{\includegraphics[width=\plotwidth,]{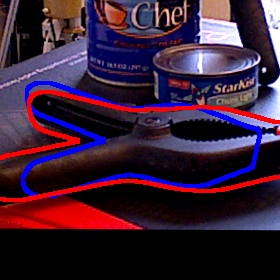}} \\

\end{tabular}
}

    \captionof{figure}{ 
     \textbf{Example results of the GoTrack refiner on YCBV~\cite{xiang2017posecnn}}.
     }
    \label{fig:ycbv}
\end{center}
\end{figure*}

\setlength\plotwidth{2.5cm}
\setlength\lineskip{1.5pt}
\setlength\tabcolsep{1.5pt} 

\begin{figure*}[!t]
\begin{center}
{\small
\begin{tabular}{
>{\centering\arraybackslash}m{\plotwidth}%
>{\centering\arraybackslash}m{\plotwidth}%
>{\centering\arraybackslash}m{\plotwidth}
>{\centering\arraybackslash}m{\plotwidth}%
>{\centering\arraybackslash}m{\plotwidth}%
>{\centering\arraybackslash}m{\plotwidth}
}

Input image & Template  & Predicted mask & Predicted flow & Warped template  & Estimated pose \\ 
\frame{\includegraphics[width=\plotwidth,]{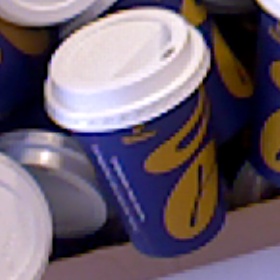}}&
\frame{\includegraphics[width=\plotwidth, ]{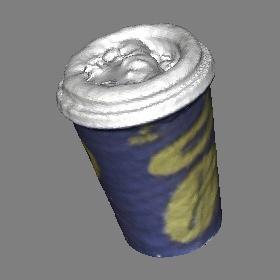}} &
\frame{\includegraphics[width=\plotwidth, ]{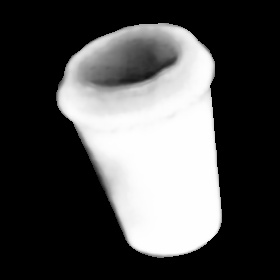}} &
\frame{\includegraphics[width=\plotwidth,]{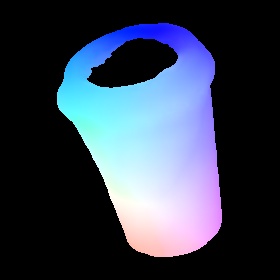}} &
\frame{\includegraphics[width=\plotwidth, ]{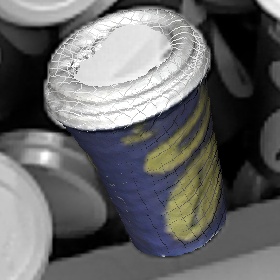}} &
\frame{\includegraphics[width=\plotwidth,]{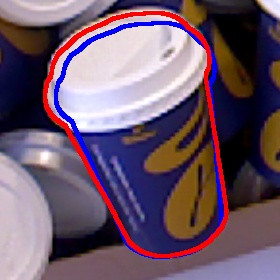}} \\

\frame{\includegraphics[width=\plotwidth,]{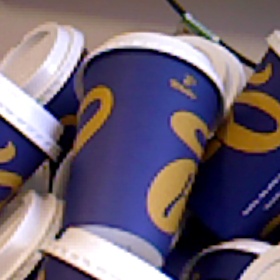}}&
\frame{\includegraphics[width=\plotwidth, ]{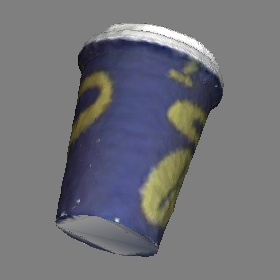}} &
\frame{\includegraphics[width=\plotwidth, ]{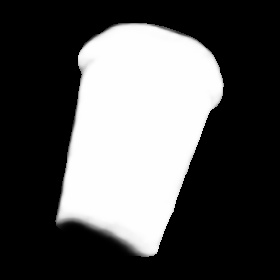}} &
\frame{\includegraphics[width=\plotwidth,]{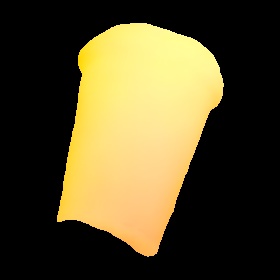}} &
\frame{\includegraphics[width=\plotwidth, ]{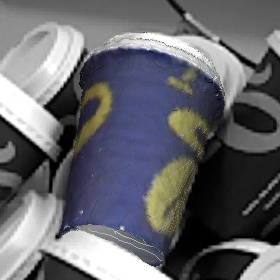}} &
\frame{\includegraphics[width=\plotwidth,]{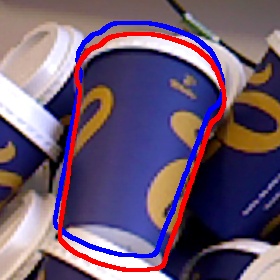}} \\

\frame{\includegraphics[width=\plotwidth,]{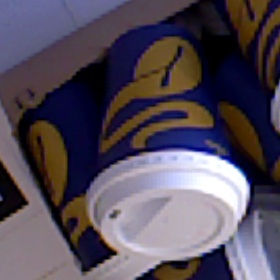}}&
\frame{\includegraphics[width=\plotwidth, ]{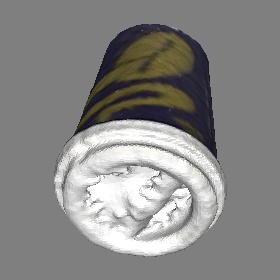}} &
\frame{\includegraphics[width=\plotwidth, ]{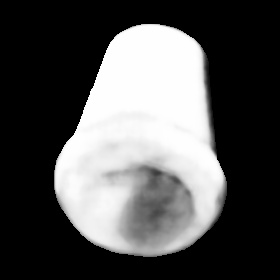}} &
\frame{\includegraphics[width=\plotwidth,]{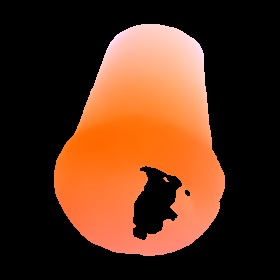}} &
\frame{\includegraphics[width=\plotwidth, ]{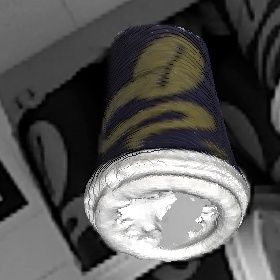}} &
\frame{\includegraphics[width=\plotwidth,]{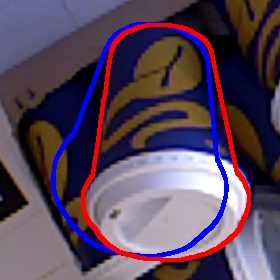}} \\

\frame{\includegraphics[width=\plotwidth,]{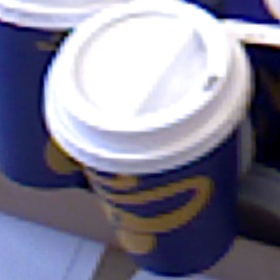}}&
\frame{\includegraphics[width=\plotwidth, ]{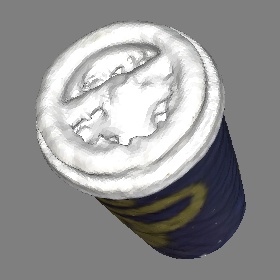}} &
\frame{\includegraphics[width=\plotwidth, ]{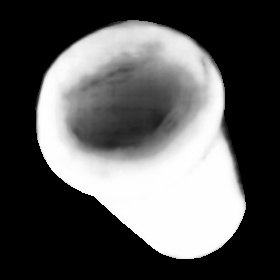}} &
\frame{\includegraphics[width=\plotwidth,]{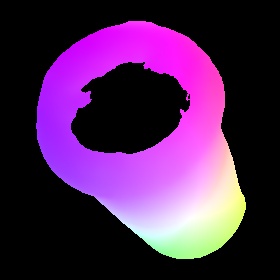}} &
\frame{\includegraphics[width=\plotwidth, ]{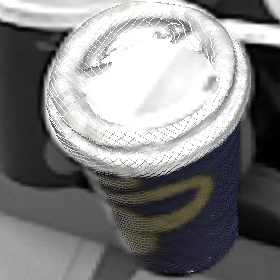}} &
\frame{\includegraphics[width=\plotwidth,]{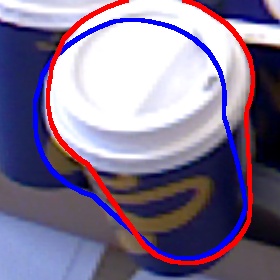}} \\

\frame{\includegraphics[width=\plotwidth,]{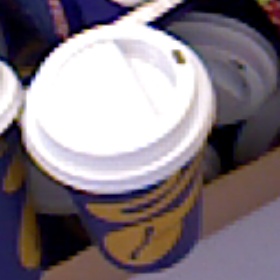}}&
\frame{\includegraphics[width=\plotwidth, ]{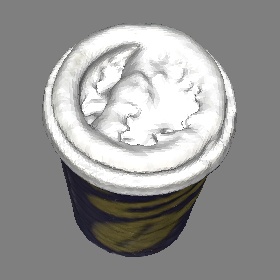}} &
\frame{\includegraphics[width=\plotwidth, ]{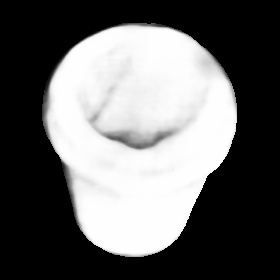}} &
\frame{\includegraphics[width=\plotwidth,]{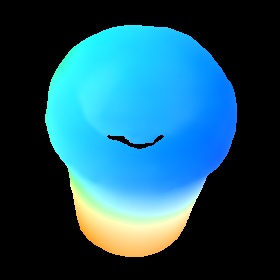}} &
\frame{\includegraphics[width=\plotwidth, ]{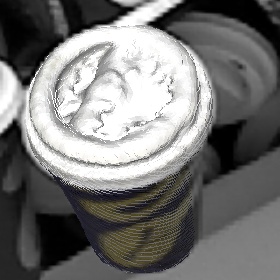}} &
\frame{\includegraphics[width=\plotwidth,]{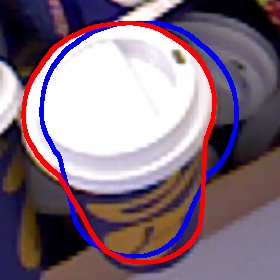}} \\

\frame{\includegraphics[width=\plotwidth,]{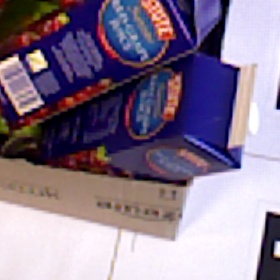}}&
\frame{\includegraphics[width=\plotwidth, ]{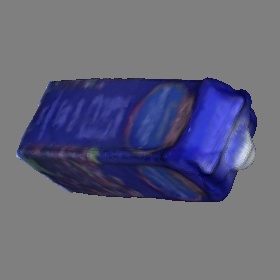}} &
\frame{\includegraphics[width=\plotwidth, ]{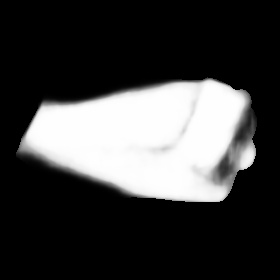}} &
\frame{\includegraphics[width=\plotwidth,]{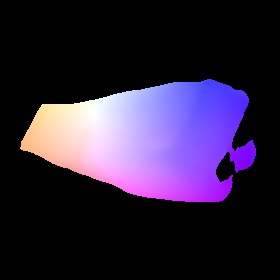}} &
\frame{\includegraphics[width=\plotwidth, ]{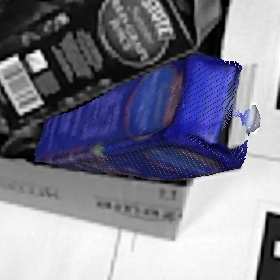}} &
\frame{\includegraphics[width=\plotwidth,]{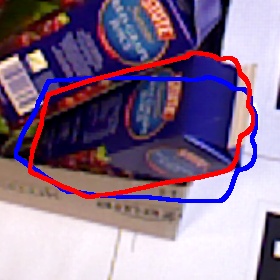}} \\

\frame{\includegraphics[width=\plotwidth,]{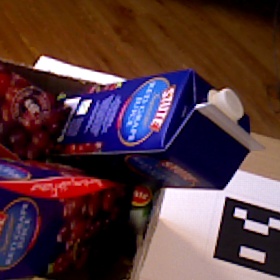}}&
\frame{\includegraphics[width=\plotwidth, ]{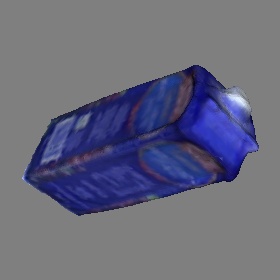}} &
\frame{\includegraphics[width=\plotwidth, ]{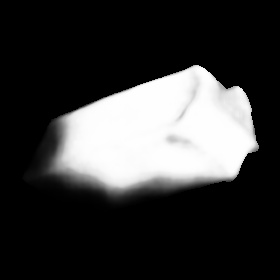}} &
\frame{\includegraphics[width=\plotwidth,]{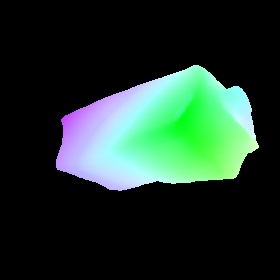}} &
\frame{\includegraphics[width=\plotwidth, ]{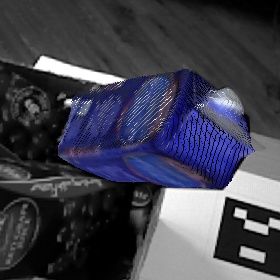}} &
\frame{\includegraphics[width=\plotwidth,]{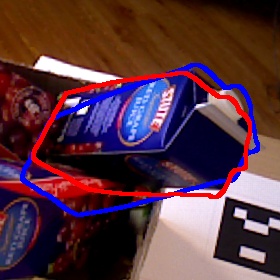}} \\

\frame{\includegraphics[width=\plotwidth,]{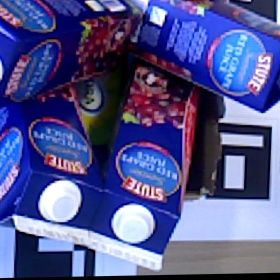}}&
\frame{\includegraphics[width=\plotwidth, ]{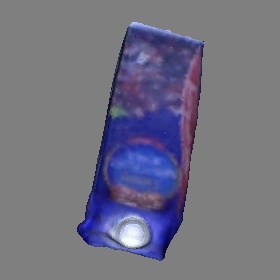}} &
\frame{\includegraphics[width=\plotwidth, ]{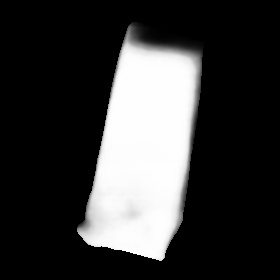}} &
\frame{\includegraphics[width=\plotwidth,]{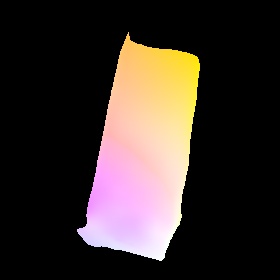}} &
\frame{\includegraphics[width=\plotwidth, ]{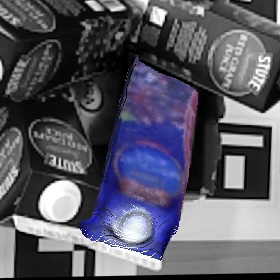}} &
\frame{\includegraphics[width=\plotwidth,]{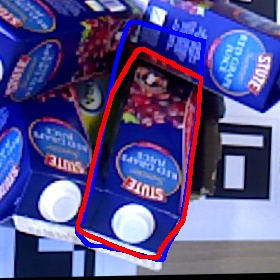}} \\

\end{tabular}
}

    \captionof{figure}{ 
     \textbf{Example results of the GoTrack refiner on IC-BIN~\cite{doumanoglou2016recovering}}.
     }
    \label{fig:icbin}
\end{center}
\end{figure*}

\setlength\plotwidth{2.5cm}
\setlength\lineskip{1.5pt}
\setlength\tabcolsep{1.5pt} 

\begin{figure*}[!t]
\begin{center}
{\small
\begin{tabular}{
>{\centering\arraybackslash}m{\plotwidth}%
>{\centering\arraybackslash}m{\plotwidth}%
>{\centering\arraybackslash}m{\plotwidth}
>{\centering\arraybackslash}m{\plotwidth}%
>{\centering\arraybackslash}m{\plotwidth}%
>{\centering\arraybackslash}m{\plotwidth}
}

Input image & Template  & Predicted mask & Predicted flow & Warped template & Estimated pose \\ 
\frame{\includegraphics[width=\plotwidth,]{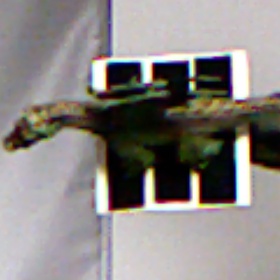}}&
\frame{\includegraphics[width=\plotwidth, ]{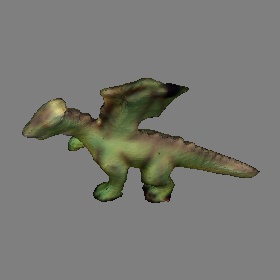}} &
\frame{\includegraphics[width=\plotwidth, ]{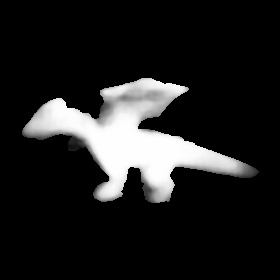}} &
\frame{\includegraphics[width=\plotwidth,]{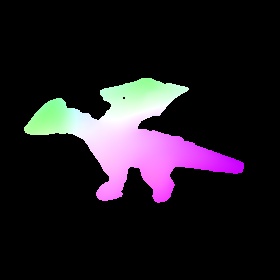}} &
\frame{\includegraphics[width=\plotwidth, ]{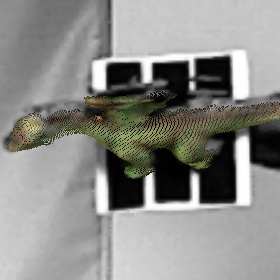}} &
\frame{\includegraphics[width=\plotwidth,]{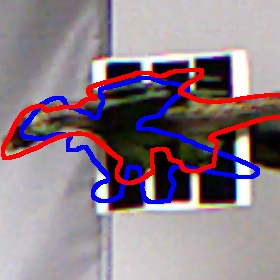}} \\

\frame{\includegraphics[width=\plotwidth,]{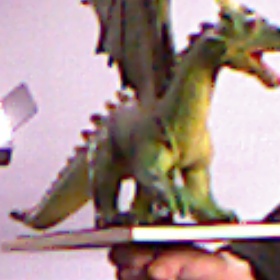}}&
\frame{\includegraphics[width=\plotwidth, ]{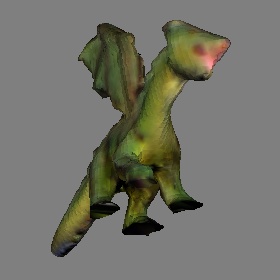}} &
\frame{\includegraphics[width=\plotwidth, ]{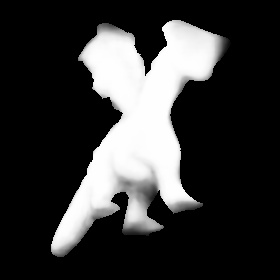}} &
\frame{\includegraphics[width=\plotwidth,]{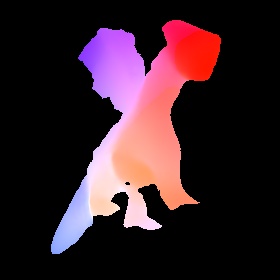}} &
\frame{\includegraphics[width=\plotwidth, ]{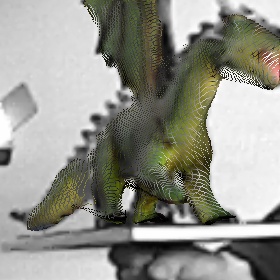}} &
\frame{\includegraphics[width=\plotwidth,]{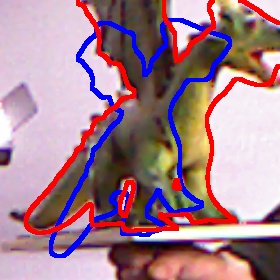}} \\

\frame{\includegraphics[width=\plotwidth,]{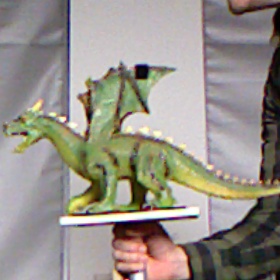}}&
\frame{\includegraphics[width=\plotwidth, ]{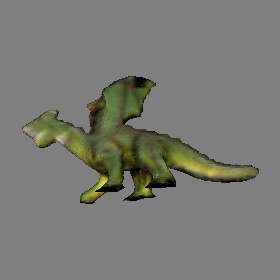}} &
\frame{\includegraphics[width=\plotwidth, ]{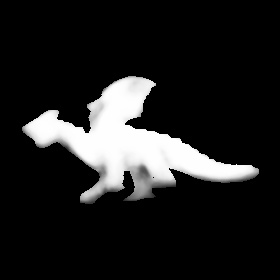}} &
\frame{\includegraphics[width=\plotwidth,]{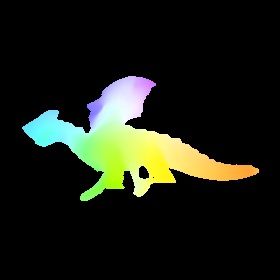}} &
\frame{\includegraphics[width=\plotwidth, ]{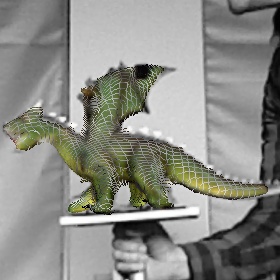}} &
\frame{\includegraphics[width=\plotwidth,]{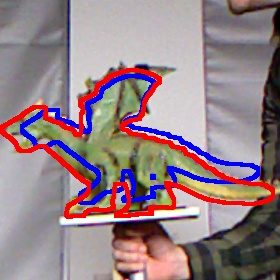}} \\

\frame{\includegraphics[width=\plotwidth,]{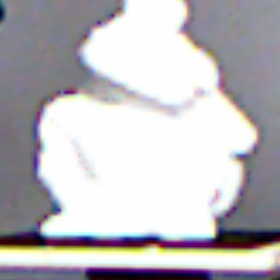}}&
\frame{\includegraphics[width=\plotwidth, ]{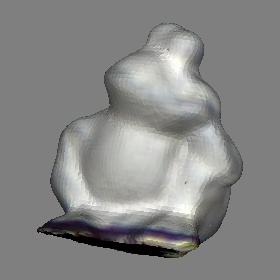}} &
\frame{\includegraphics[width=\plotwidth, ]{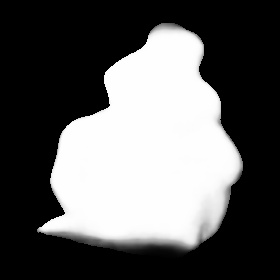}} &
\frame{\includegraphics[width=\plotwidth,]{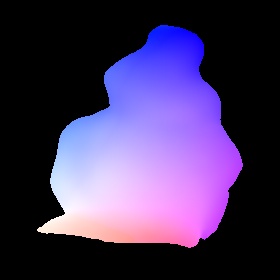}} &
\frame{\includegraphics[width=\plotwidth, ]{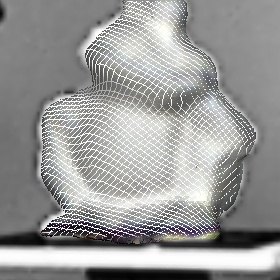}} &
\frame{\includegraphics[width=\plotwidth,]{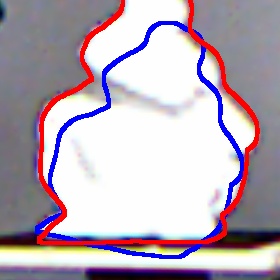}} \\

\frame{\includegraphics[width=\plotwidth,]{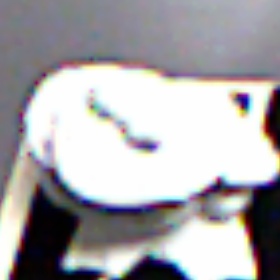}}&
\frame{\includegraphics[width=\plotwidth, ]{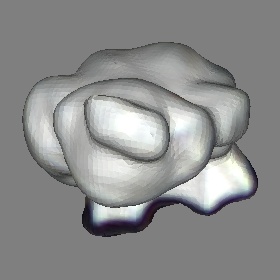}} &
\frame{\includegraphics[width=\plotwidth, ]{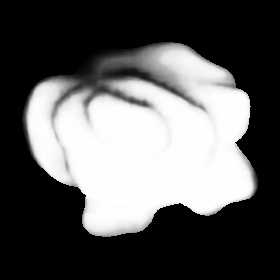}} &
\frame{\includegraphics[width=\plotwidth,]{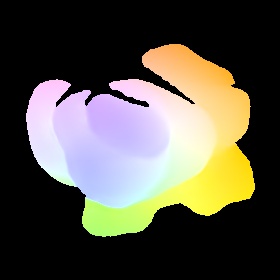}} &
\frame{\includegraphics[width=\plotwidth, ]{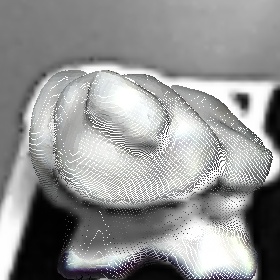}} &
\frame{\includegraphics[width=\plotwidth,]{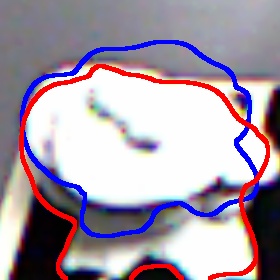}} \\

\frame{\includegraphics[width=\plotwidth,]{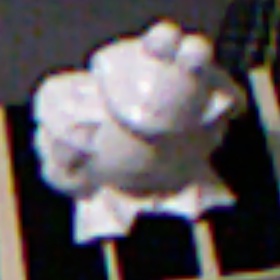}}&
\frame{\includegraphics[width=\plotwidth, ]{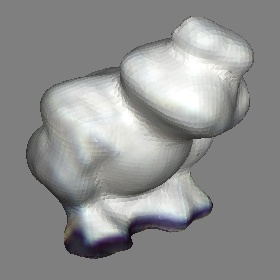}} &
\frame{\includegraphics[width=\plotwidth, ]{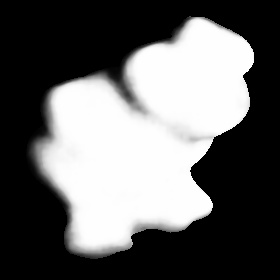}} &
\frame{\includegraphics[width=\plotwidth,]{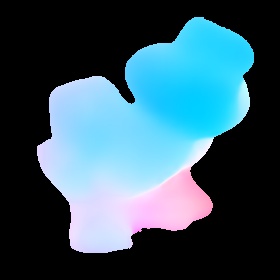}} &
\frame{\includegraphics[width=\plotwidth, ]{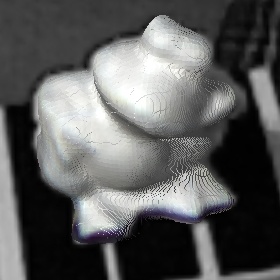}} &
\frame{\includegraphics[width=\plotwidth,]{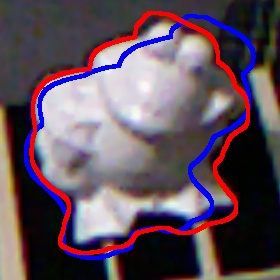}} \\

\frame{\includegraphics[width=\plotwidth,]{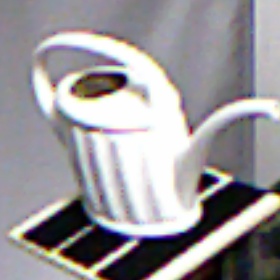}}&
\frame{\includegraphics[width=\plotwidth, ]{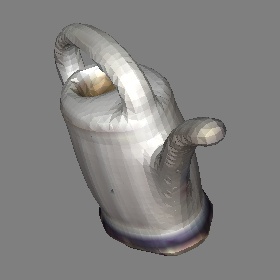}} &
\frame{\includegraphics[width=\plotwidth, ]{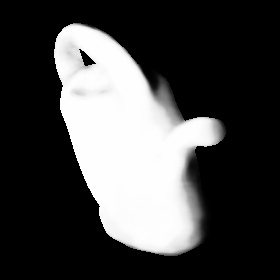}} &
\frame{\includegraphics[width=\plotwidth,]{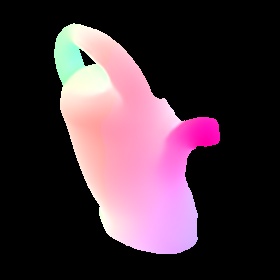}} &
\frame{\includegraphics[width=\plotwidth, ]{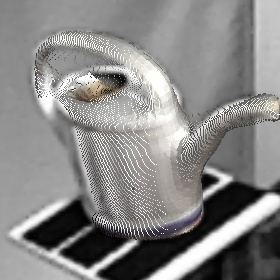}} &
\frame{\includegraphics[width=\plotwidth,]{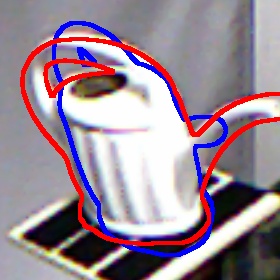}} \\

\frame{\includegraphics[width=\plotwidth,]{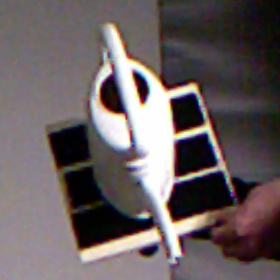}}&
\frame{\includegraphics[width=\plotwidth, ]{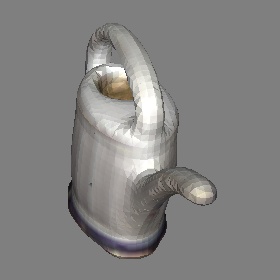}} &
\frame{\includegraphics[width=\plotwidth, ]{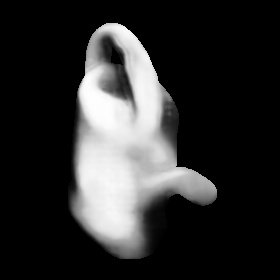}} &
\frame{\includegraphics[width=\plotwidth,]{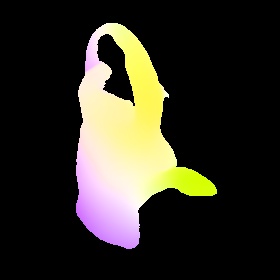}} &
\frame{\includegraphics[width=\plotwidth, ]{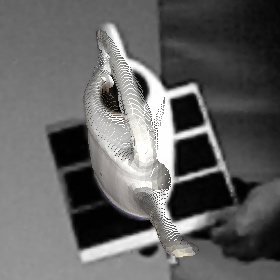}} &
\frame{\includegraphics[width=\plotwidth,]{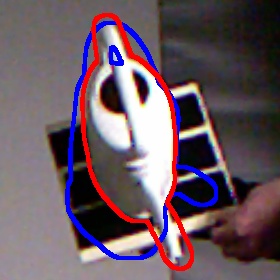}} \\

\end{tabular}
}

    \captionof{figure}{ 
     \textbf{Example results of the GoTrack refiner on TUDL~\cite{hodan2018bop}}.
     }
    \label{fig:tudl}
\end{center}
\end{figure*}

\setlength\plotwidth{2.5cm}
\setlength\lineskip{1.5pt}
\setlength\tabcolsep{1.5pt} 

\begin{figure*}[!t]
\begin{center}
{\small
\begin{tabular}{
>{\centering\arraybackslash}m{\plotwidth}%
>{\centering\arraybackslash}m{\plotwidth}%
>{\centering\arraybackslash}m{\plotwidth}
>{\centering\arraybackslash}m{\plotwidth}%
>{\centering\arraybackslash}m{\plotwidth}%
>{\centering\arraybackslash}m{\plotwidth}
>{\centering\arraybackslash}m{\plotwidth}
}

Input image & Template  & Predicted mask & Predicted flow & Warped template & Estimated pose \\ 
\frame{\includegraphics[width=\plotwidth,]{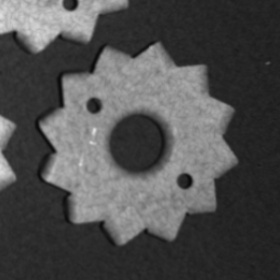}}&
\frame{\includegraphics[width=\plotwidth, ]{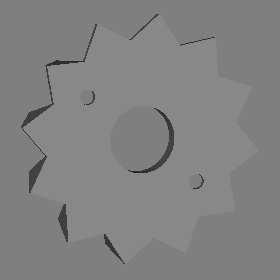}} &
\frame{\includegraphics[width=\plotwidth, ]{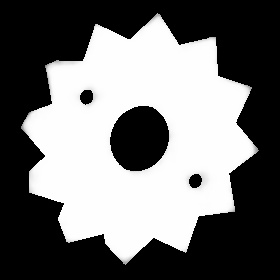}} &
\frame{\includegraphics[width=\plotwidth,]{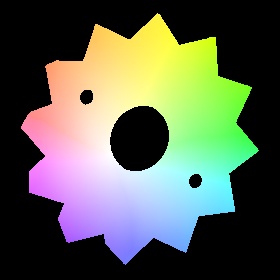}} &
\frame{\includegraphics[width=\plotwidth, ]{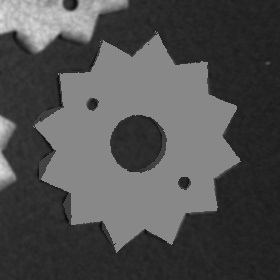}} &
\frame{\includegraphics[width=\plotwidth,]{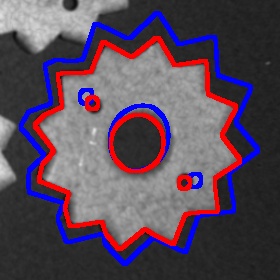}} \\

\frame{\includegraphics[width=\plotwidth,]{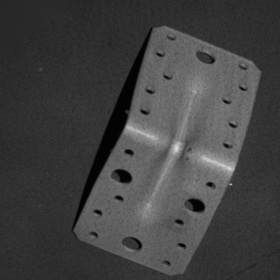}}&
\frame{\includegraphics[width=\plotwidth, ]{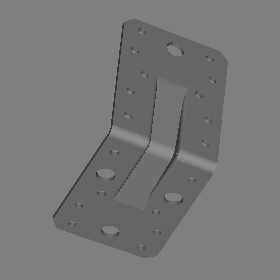}} &
\frame{\includegraphics[width=\plotwidth, ]{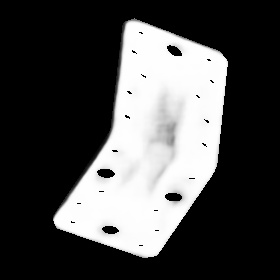}} &
\frame{\includegraphics[width=\plotwidth,]{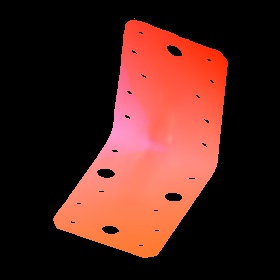}} &
\frame{\includegraphics[width=\plotwidth, ]{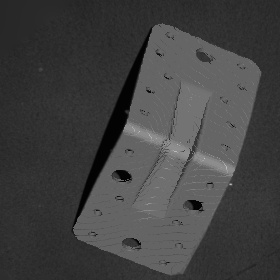}} &
\frame{\includegraphics[width=\plotwidth,]{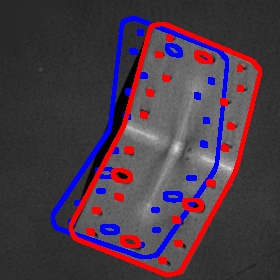}} \\

\frame{\includegraphics[width=\plotwidth,]{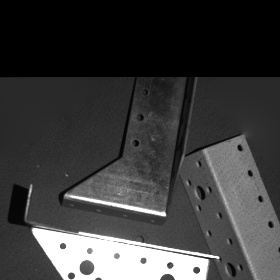}}&
\frame{\includegraphics[width=\plotwidth, ]{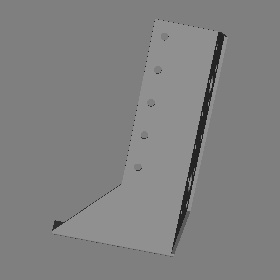}} &
\frame{\includegraphics[width=\plotwidth, ]{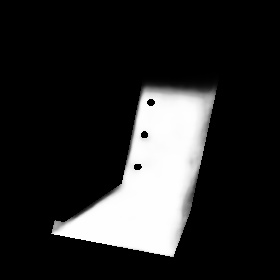}} &
\frame{\includegraphics[width=\plotwidth,]{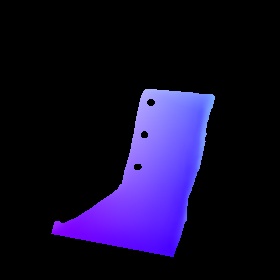}} &
\frame{\includegraphics[width=\plotwidth, ]{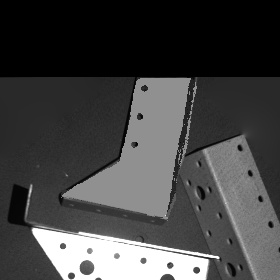}} &
\frame{\includegraphics[width=\plotwidth,]{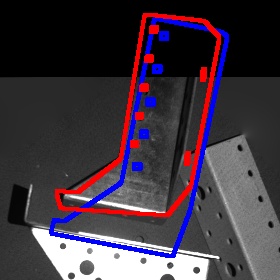}} \\

\frame{\includegraphics[width=\plotwidth,]{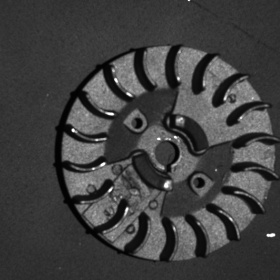}}&
\frame{\includegraphics[width=\plotwidth, ]{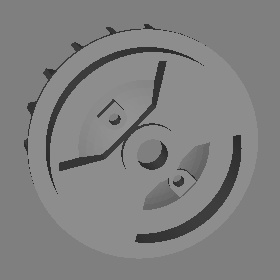}} &
\frame{\includegraphics[width=\plotwidth, ]{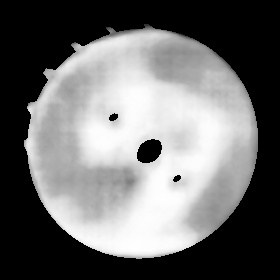}} &
\frame{\includegraphics[width=\plotwidth,]{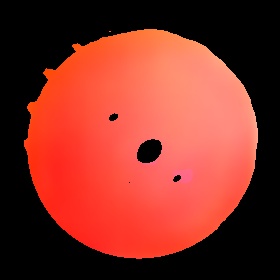}} &
\frame{\includegraphics[width=\plotwidth, ]{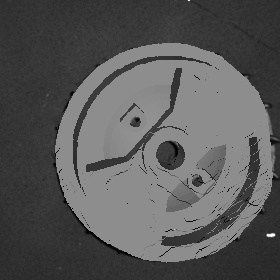}} &
\frame{\includegraphics[width=\plotwidth,]{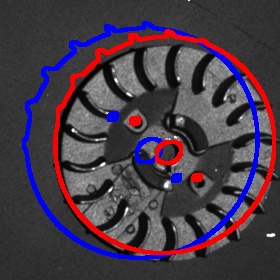}} \\

\frame{\includegraphics[width=\plotwidth,]{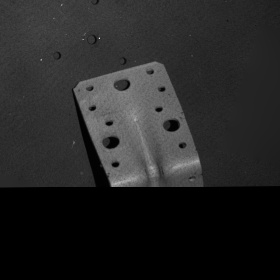}}&
\frame{\includegraphics[width=\plotwidth, ]{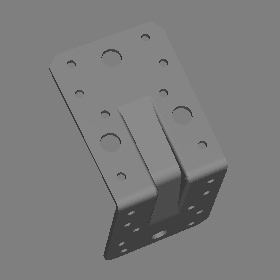}} &
\frame{\includegraphics[width=\plotwidth, ]{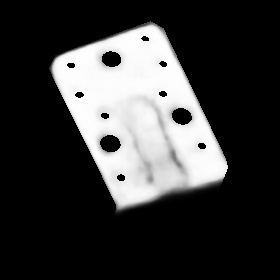}} &
\frame{\includegraphics[width=\plotwidth,]{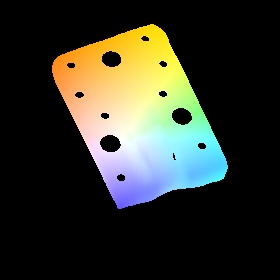}} &
\frame{\includegraphics[width=\plotwidth, ]{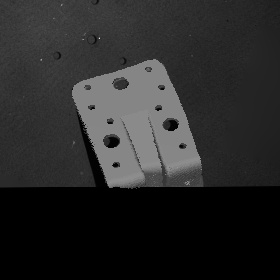}} &
\frame{\includegraphics[width=\plotwidth,]{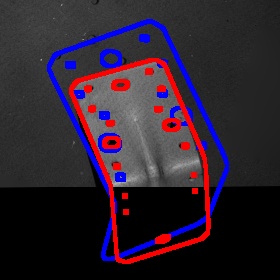}} \\
\frame{\includegraphics[width=\plotwidth,]{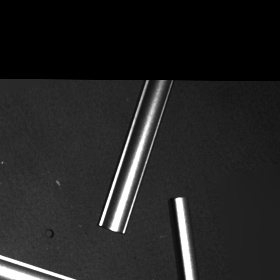}}&
\frame{\includegraphics[width=\plotwidth, ]{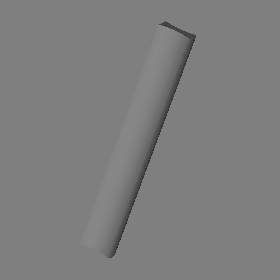}} &
\frame{\includegraphics[width=\plotwidth, ]{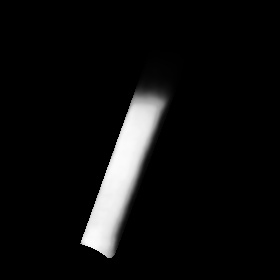}} &
\frame{\includegraphics[width=\plotwidth,]{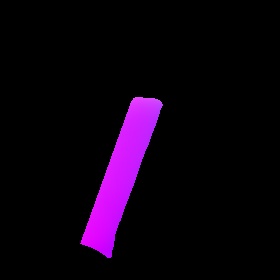}} &
\frame{\includegraphics[width=\plotwidth, ]{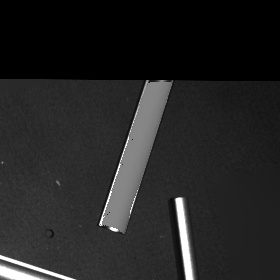}} &
\frame{\includegraphics[width=\plotwidth,]{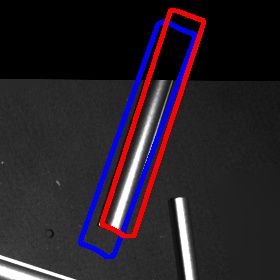}} \\

\frame{\includegraphics[width=\plotwidth,]{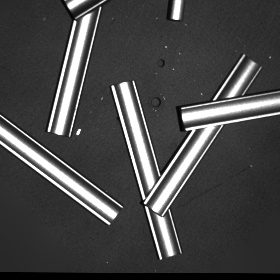}}&
\frame{\includegraphics[width=\plotwidth, ]{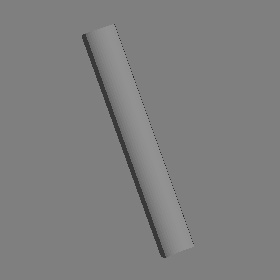}} &
\frame{\includegraphics[width=\plotwidth, ]{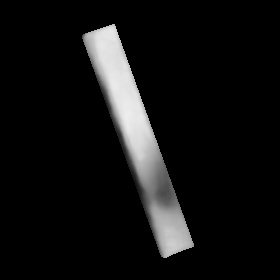}} &
\frame{\includegraphics[width=\plotwidth,]{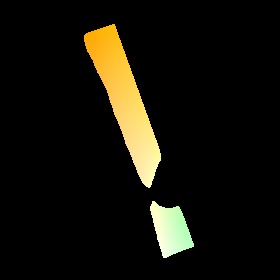}} &
\frame{\includegraphics[width=\plotwidth, ]{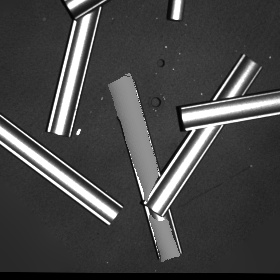}} &
\frame{\includegraphics[width=\plotwidth,]{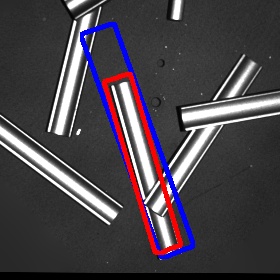}} \\

\frame{\includegraphics[width=\plotwidth,]{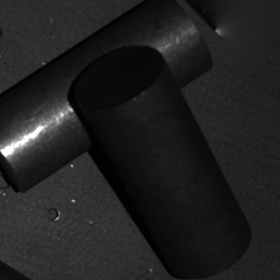}}&
\frame{\includegraphics[width=\plotwidth, ]{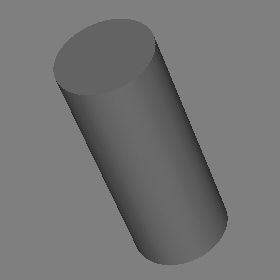}} &
\frame{\includegraphics[width=\plotwidth, ]{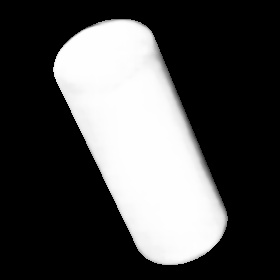}} &
\frame{\includegraphics[width=\plotwidth,]{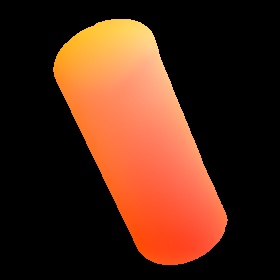}} &
\frame{\includegraphics[width=\plotwidth, ]{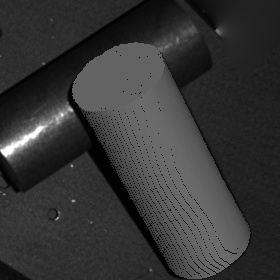}} &
\frame{\includegraphics[width=\plotwidth,]{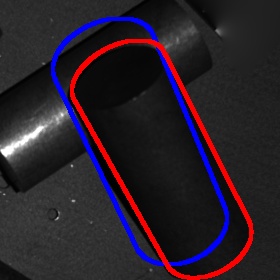}} \\

\end{tabular}
}

    \captionof{figure}{ 
     \textbf{Example results of the GoTrack refiner on ITODD~\cite{drost2017introducing}}.
     }
    \label{fig:itodd}
\end{center}
\end{figure*}

\end{document}